\documentclass[letterpaper]{article}

\usepackage[preprint]{aaai2027}
\usepackage[hyphens]{url}
\usepackage{graphicx}
\usepackage{natbib}
\usepackage{caption}
\usepackage{booktabs}
\usepackage{amsmath,amssymb,mathtools,bm}
\usepackage{microtype}
\usepackage{multirow}
\usepackage{array}
\usepackage{tabularx}
\usepackage{makecell}
\usepackage{algorithm}
\usepackage{algpseudocode}
\title{PhysMAS: Physics-Grounded Multi-Agent Synthesis\\of Compositional 4D Gaussians}

\author{
Jiang Qin\textsuperscript{\rm 1}\equalcontrib,
Chunji Lv\textsuperscript{\rm 1}\equalcontrib,
Yangguang Wei\textsuperscript{\rm 2},
Yang Gao\textsuperscript{\rm 2},
Ming Liu\textsuperscript{\rm 2},\\
Lizhong Ding\textsuperscript{\rm 1},
Ye Yuan\textsuperscript{\rm 1},
Yinjie Lei\textsuperscript{\rm 3},
Changsheng Li\textsuperscript{\rm 1}\corresponding
}

\affiliations{
\textsuperscript{\rm 1}Beijing Institute of Technology\\

\textsuperscript{\rm 2}Meituan\\

\textsuperscript{\rm 3}Sichuan University
}
\newcommand{\method}{PhysMAS}
\newcommand{\ours}{\method{} (Ours)}
\newcommand{\particleset}{\mathcal{Q}}

\begin{document}

\maketitle

\begin{abstract}
Efficient, fully automatic, and physically plausible 4D Gaussian synthesis is an important goal for dynamic scene generation. Recent physics-based methods couple 3D Gaussians with the Material Point Method (MPM) to generate physically driven motion, but extending this paradigm to heterogeneous multi-part objects and interacting multi-object scenes remains challenging. Object-level physical assignment collapses distinct parts into a single material state, while one-shot predictions from large language models, vision-language models, or agents neither reliably bind different materials to identified parts nor verify that the resulting MPM configuration is executable. Score Distillation Sampling (SDS)-based parameter optimization, meanwhile, requires repeated per-scene score evaluations and gradient backpropagation, incurring lengthy optimization and potentially yielding suboptimal or unstable solutions. We therefore present PhysMAS, a physics-grounded multi-agent framework. From a motion prompt and four scene views, an Object-Part Scene Agent establishes persistent identities and calls a Material Reasoning Agent for part-wise profiles. It invokes solver-aware skills to bind these identities and profiles to per-particle MPM fields and execute all objects in a shared domain; the framework then screens candidate forward-simulation results. This supports heterogeneous multi-part and interacting multi-object scenes without per-scene diffusion-score backpropagation. Extensive experiments demonstrate that, compared with recent physics-based 4D Gaussian baselines that rely on SDS, PhysMAS achieves better semantic alignment and perceived physical plausibility while requiring less runtime.
\end{abstract}

\section{Introduction}
\label{sec:intro}

\begin{figure*}[t]
\centering
\includegraphics[width=\textwidth]{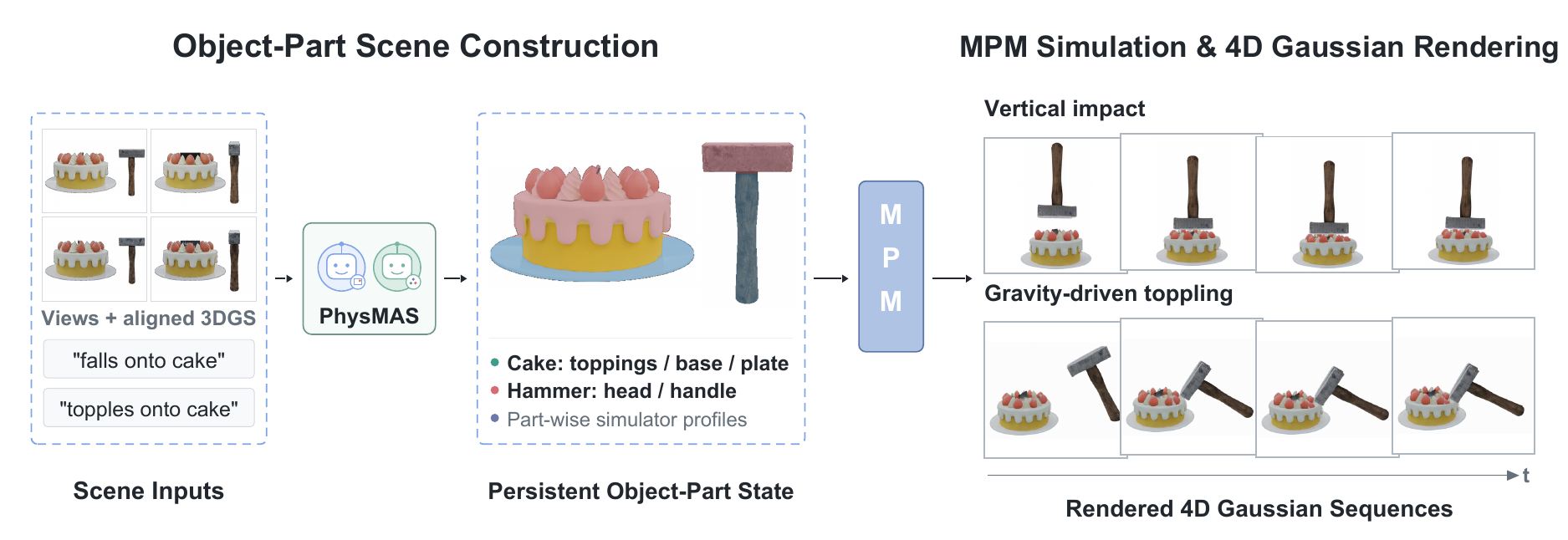}
\caption{Overview of PhysMAS. Each run receives a motion prompt, four calibrated RGB views, their aligned static 3DGS, and prescribed output cameras. PhysMAS constructs persistent object--part state with part-wise simulator profiles; the compiled state is advanced in a shared MPM domain and rendered as a 4D Gaussian sequence. Shown are two runs of the same registered scene bundle: vertical impact and gravity-driven toppling.}
\label{fig:teaser}
\end{figure*}

Efficient simulator-driven 4D Gaussian synthesis is an important goal in computer graphics and generative AI. 3D Gaussian Splatting (3DGS) provides an explicit representation for high-quality rendering \citep{kerbl2023gaussian}, and physics-based methods couple its primitives with the Material Point Method (MPM) \citep{sulsky1994particle,stomakhin2013snow} to generate physically driven motion \citep{xie2024physgaussian}. Yet many physics-based 4D Gaussian pipelines assign properties at the object or Gaussian-field level without an explicit hierarchy linking parts to their parent objects across perception and simulation. This is restrictive when objects interact or contain parts with different structural roles and material responses. In this work, \emph{compositional} refers to scene state indexed jointly by object and part. Figure~\ref{fig:teaser} illustrates the resulting dynamics.

This extension exposes two coupled problems. First, without persistent object--part identity, visual evidence cannot be bound consistently to the corresponding part and object. Object-wide assignment forces heterogeneous parts to share one material state, while co-occurring objects must remain distinct inside a shared simulation. Second, an LLM or VLM can propose labels and parameters, but a one-time prediction does not enforce the profiles and numerical constraints accepted by MPM. Alternatives based on Score Distillation Sampling (SDS) \citep{poole2023dreamfusion} optimize parameters with video priors, yet require repeated per-scene score evaluation and backpropagation \citep{huang2025dreamphysics,lin2025omniphysgs}. A practical system therefore requires compositional scene state and simulator-constrained execution without per-scene diffusion-score optimization.

We propose \textbf{PhysMAS}, a physics-grounded multi-agent framework that receives a motion prompt, four calibrated RGB views with their aligned static 3DGS, and prescribed output cameras. An \textbf{Object-Part Scene Agent} grounds object instances and semantic parts, assigns compound object--part labels to the scene Gaussians, and calls a \textbf{Material Reasoning Agent} for part-wise simulator profiles and physical parameters. Deterministic, solver-aware \textbf{MPM skills} transfer these labels and profiles to particles, execute all objects in one domain, and return execution records for validity screening. Within a shared single-velocity grid, standard MPM updates coordinate momentum exchange among objects and support the multi-object interactions evaluated in our experiments. Thus, PhysMAS automates object--part grounding, material profiling, MPM compilation, simulation, and screening without per-scene SDS backpropagation.

In summary, our main contributions are as follows:
\begin{itemize}
    \item We introduce persistent object--part state for representing heterogeneous parts and multiple objects in an aligned static 3DGS using four calibrated scene views.
    \item We formalize an auditable two-agent compiler contract around this typed state: the scene agent owns identity, planning, and acceptance; the material agent returns row-wise profile evidence; deterministic routines compile, execute, and screen part-wise candidates.
    \item Across the multi-part and multi-object evaluations, PhysMAS obtains the highest observed mean frame-wise CLIP$_{sim}$ and the lowest paired-reference image-plane trajectory, projected-area, and local-strain errors among the compared methods; it achieves a two-minute reported runtime and is substantially faster than the evaluated long per-scene SDS optimization baselines.
\end{itemize}

\section{Related Work}
\label{sec:related}

\subsection{Generative 4D Gaussian Splatting}

Dynamic Gaussian representations include 4D-GS, Deformable 3D Gaussians, Spacetime Gaussians, and L4GM \citep{wu2024fourDGS,yang2024deformablegs,li2024spacetime,ren2024l4gm}. Diffusion-guided 4D generation builds on DreamFusion and includes Align Your Gaussians, Animate3D, 4Diffusion, and CAT4D \citep{poole2023dreamfusion,ling2024align,jiang2024animate3d,zhang2024fourdiffusion,wu2025cat4d}. They target reconstruction or plausible animation rather than material-conditioned object--part simulation.

\subsection{Physics-Grounded 4D Gaussian Splatting}

Physics-grounded methods connect Gaussian appearance to MPM \citep{sulsky1994particle,stomakhin2013snow}; PhysGaussian provides the bridge, while Gaussian Splashing studies rotation artifacts \citep{xie2024physgaussian,feng2025gaussiansplashing}. PhysDreamer, DreamPhysics, MotionPhysics, OmniPhysGS, PhysSplat, and PhysGM automate property inference \citep{zhang2024physdreamer,huang2025dreamphysics,wang2026motionphysics,lin2025omniphysgs,zhao2025physsplat,lv2025physgm}. Feature Splatting, GaussianProperty, PhysGS, and GaussianFluent address semantic or physical attributes and mixed materials \citep{qiu2024featuresplatting,xu2025gaussianproperty,chopra2026physgs,huang2026gaussianfluent}. PAC-NeRF, GIC, Vid2Sim, FreeGave, and PhysTwin infer dynamics, properties, or simulation-ready representations from observations \citep{li2023pacnerf,cai2024gic,chen2025vid2sim,li2025freegave,jiang2025phystwin}. PhysMAS introduces an object-scoped part table that remains addressable through profile proposal, particle compilation, and executable candidate screening.

\subsection{Object-Part Scene Grounding and Agentic Physical Reasoning}

PartSLIP and SAGA address open-vocabulary or promptable 3D segmentation \citep{liu2023partslip,cen2025saga}; Gaussian Grouping and LangSplat attach identity or language semantics to Gaussians \citep{ye2024gaussiangrouping,qin2024langsplat}; Part$^2$GS recovers part structure \citep{yu2026part2gs}. Physics-scene methods cover object worlds, contact, heterogeneous solvers, and multi-body evaluation \citep{liu2024physgen,chen2025physgen3d,wang2025decoupledgaussian,jiang2026physho,li2025trace,liu2026mosiv,kim2026physgaia}. Language agents support programmatic interaction with environments \citep{ma2024eureka}; PhysAgent refines simulator force fields in the loop \citep{lv2026physagent}. PhysMAS maintains typed object-scoped part state from grounding through material candidates, particle compilation, and screening; its two-agent contract assigns identity, planning, and acceptance to the scene agent, while the material agent supplies row-wise evidence.

\section{Method}
\label{sec:method}

\begin{figure*}[!t]
\centering
\includegraphics[width=\textwidth]{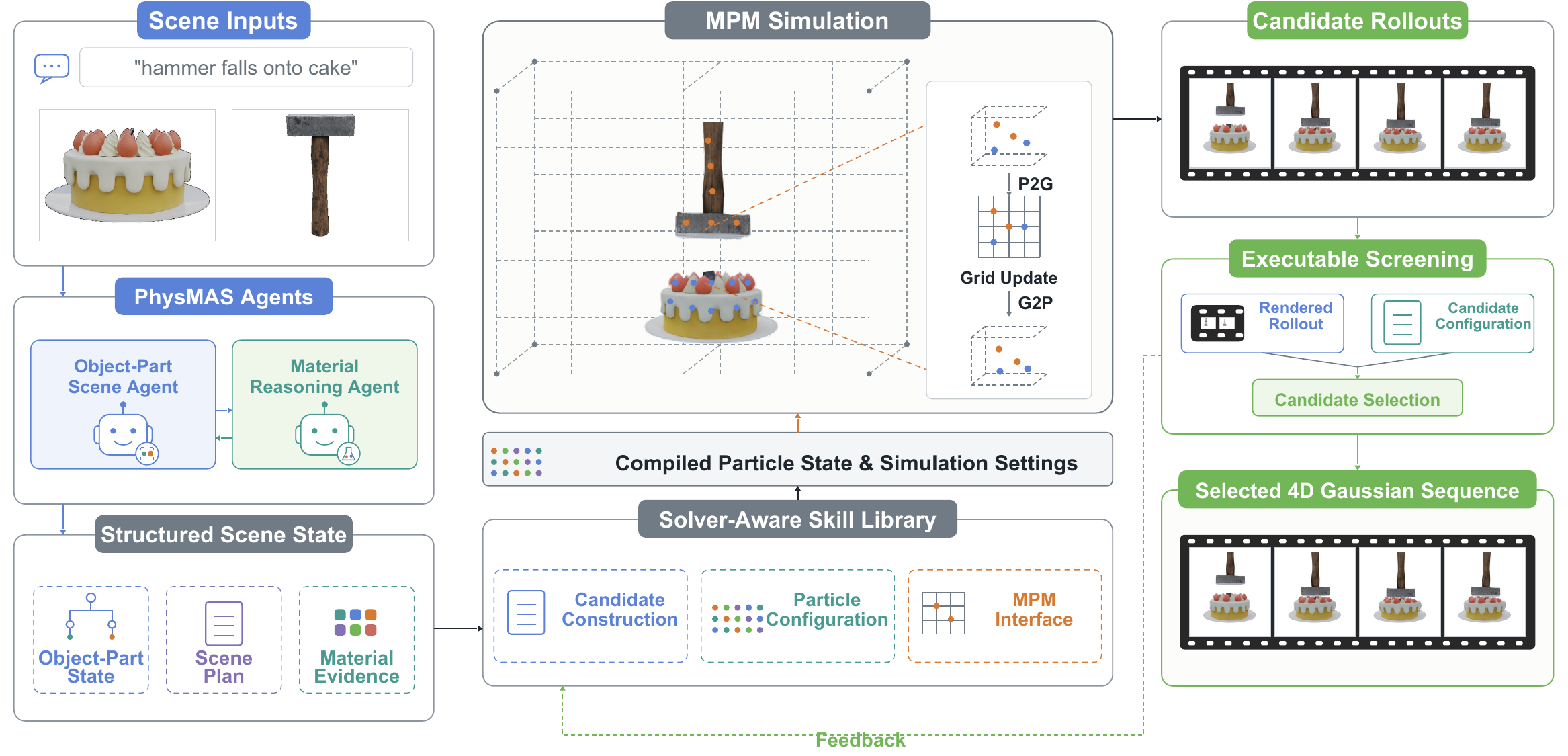}
\caption{PhysMAS framework. The agents transform a motion prompt, four calibrated views, and an aligned static 3DGS into object--part state, a scene plan, and material evidence. Solver-aware skills compile and simulate candidates, screen their execution records, and return the accepted 4D Gaussian sequence.}
\label{fig:overview}
\end{figure*}

PhysMAS receives a motion prompt $y$, four calibrated RGB views $\mathcal I$ with an aligned 3DGS $\mathcal G^0$, and prescribed output cameras. The scene agent maintains object--part identity, supported plans, and acceptance; the material agent supplies row-wise profiles; deterministic skills compile, simulate, screen, and repair candidates (Figure~\ref{fig:overview}).

\subsection{Problem Setup}
\label{sec:formulation}

Let $\mathcal I=\{(I_v,\mathsf C_v)\}_{v=1}^{4}$ be the ordered RGB views and calibrated cameras, and $\mathcal G^0=\{g_i^0\}_{i=1}^{N_G}$ the aligned scene-level 3DGS, with $g_i^0=(\bm\mu_i^0,\bm\Sigma_i^0,\alpha_i,\bm h_i)$. A fixed run specification provides bounded gravity/colliders, supported velocity or position blends, and impulses. Given $y$ and output cameras $\{\mathsf C_{\mathrm{out}}^t\}_{t=1}^{T}$, PhysMAS advances constitutive MPM under compiled controls and returns $\mathcal V=\{\operatorname{Render}(\mathcal G^t,\mathsf C_{\mathrm{out}}^t)\}_{t=1}^{T}$. Each Gaussian carries an object identity and object-scoped part row $r=(o,p)$; semantic and residual rows form the active set $\mathcal R^{\mathrm{act}}$ indexing material state. Identically named parts of different objects remain distinct, while a one-part object uses one semantic row.

\subsection{Object-Part Scene Agent}
\label{sec:objectpart}

The backbone proposes a serialized object--part inventory whose object IDs serve as cross-view query indices. GroundingDINO and SAM~2 \citep{liu2024groundingdino,ravi2024sam2} provide masks and confidences; after Gaussian assignment, the compound labels remain fixed through material proposal, particle compilation, and rendering. From the prompt and inventory, the agent forms scene plan $\Pi$ using supported roles, action targets, directions, magnitudes, and active intervals. The fixed run specification supplies gravity and colliders, while a deterministic compiler canonicalizes and validates the plan. The run specification also stores asset geometry, $\mathcal G^0$, output cameras, grid resolution, timing, step caps, and rendering settings; $\Pi$ contains the structured scene actions. Let $\mathcal O$ be the detected objects and $\mathcal R_o$ the rows of object $o$. Perception is object-first: object evidence is aggregated over camera-valid projections, then part evidence is compared only within $\mathcal R_o$. If $\Omega_i$ contains views where Gaussian $i$ has positive depth and projects inside the image at $\bm u_{iv}$, object assignment is
\begin{equation}
b_{io}=\sum_{v\in\Omega_i}e^{\mathrm{obj}}_{ov}(\bm u_{iv}),
\qquad
\widetilde o_i=\operatorname*{arg\,max}_{o\in\mathcal O}b_{io}.
\label{eq:objectscore}
\end{equation}
Conditioned on $\widetilde o_i$, the within-object part assignment is
\begin{equation}
a_{ir}=\sum_{v\in\Omega_i}e^{\mathrm{part}}_{rv}(\bm u_{iv}),
\qquad
\widetilde z_i^G=\operatorname*{arg\,max}_{r\in\mathcal R_{\widetilde o_i}}a_{ir}.
\label{eq:partscore}
\end{equation}
The evidence functions combine mask confidence and spatial support. Deterministic propagation fuses $(\widetilde o_i,\widetilde z_i^G)$ into $z_i^G$ and adds object-scoped residual rows for complete coverage. The resulting $\bm z^G$ is fixed before material reasoning; guards and ties are specified in the supplement.

\subsection{Material Reasoning Agent}
\label{sec:material}

For each $r\in\mathcal R^{\mathrm{act}}$, evidence $\mathcal H_r$ contains highlighted observations, its name, and structural role. The Material Reasoning Agent combines vision--language material/branch scores with $E$/$\nu$ statistics from a frozen crop-conditioned predictor \citep{lv2025physgm} to return $\bm\psi_r=\mathcal A_{\mathrm{mat}}(\mathcal H_r)$. After row-wise validation, candidate records $\Theta_r^{(c)}=(\beta_r^{(c)},E_r^{(c)},\nu_r^{(c)},\rho_r^{(c)})$ are constrained to the accepted simulator-profile domain $\mathcal K_{\mathrm{sim}}$ (Jelly, Metal, Sand, Foam, Snow, and Plasticine). Candidate construction yields a part table and controls
\begin{equation}
\bigl(\{\Theta_r^{(c)}\}_r,\mathcal U_{\mathrm{aux}}^{(c)}\bigr)
 =\mathcal T_c\bigl(\{\bm\psi_r\}_r;\mathcal K_{\mathrm{sim}},\Pi\bigr).
\label{eq:candidateconstruction}
\end{equation}
The fixed scene state $\mathcal Z^0$ contains $\mathcal G^0$, Gaussian labels, initialized particles, numerical settings, and output cameras. The compiler then forms the schedule and simulator input
\begin{equation}
\begin{aligned}
\Gamma^{(c)}&=\operatorname{Compile}(\Pi,\mathcal U_{\mathrm{aux}}^{(c)}),\\
\mathcal C^{(c)}&=\bigl(\mathcal Z^0,\{\Theta_r^{(c)}\}_r,\Gamma^{(c)}\bigr).
\end{aligned}
\label{eq:materialstate}
\end{equation}
Here $\beta$ is the constitutive branch, while $E$, $\nu$, and $\rho$ are Young's modulus, Poisson's ratio, and density; profile-specific yield, hardening, and return-map constants remain fixed. Starting from posterior medians, fixed transforms form MAP, target-impact, support-bonded/stiff, solver-compatible, projection, soft/stiff, uncertainty-quantile, and second-profile variants; applicable candidates are retained in a fixed order. Their transforms and budgets are specified in the supplement. Every candidate preserves $\mathcal Z^0$, identities, and scene actions $\Pi$; $\mathcal U_{\mathrm{aux}}^{(c)}$ contains only structure/interface controls.

\subsection{Solver-Aware Shared-Domain Simulation}
\label{sec:skills}

Let $\mathcal Q=\{1,\ldots,N_Q\}$ index the particles in $\mathcal Z^0$. The map $\mathcal N(\bm x)=\bm L\bm x+\bm d$, with isotropic $\bm L=s\bm I_3$ and $s>0$, converts scene coordinates to normalized MPM space. The first $N_G$ particles are Gaussian means $\widehat{\bm\mu}_i^0=\mathcal N(\bm\mu_i^0)$; support-filling particles retain source-object identity $o(q)$ and source Gaussian
\begin{equation}
\iota(q)=
\begin{cases}
q,&q\leq N_G,\\
\operatorname*{arg\,min}_{i:\,o_i=o(q)}
\|\bm x_q^0-\widehat{\bm\mu}_i^0\|_2,&q>N_G,
\end{cases}
\label{eq:particlesource}
\end{equation}
The candidate is then bound to per-particle identity, parameters, and mass:
\begin{equation}
\begin{aligned}
z_q^Q&=z_{\iota(q)}^G,\qquad
\vartheta_q^{(c)}=\operatorname{Bind}(\Theta_{z_q^Q}^{(c)}),\\
M_q^{(c)}&=\rho_q^{(c)}V_q^0.
\end{aligned}
\label{eq:particlearray}
\end{equation}
Here $V_q^0$ is rest volume and $\operatorname{Bind}$ builds the branch record; residual rows ensure complete particle coverage. Affine Particle-In-Cell (APIC) transfers advance the MPM state \citep{sulsky1994particle,stomakhin2013snow,jiang2015apic,xie2024physgaussian}, with parameters, gravity, and substeps defined in the coordinates induced by $\mathcal N$.

Omitting candidate superscript $(c)$, APIC particle-to-grid (P2G) transfer at node $a$ accumulates mass and momentum from every object:
\begin{equation}
\begin{aligned}
m_a&=\sum_q w_{aq}M_q,\\
\bm p_a
&=\sum_qw_{aq}M_q
\!\left[\bm v_q+\bm B_q(\bm x_a-\bm x_q)\right]\\
&\quad+\Delta t\,\bm f_a^{\mathrm{int}}.
\end{aligned}
\label{eq:p2g}
\end{equation}
The provisional grid velocity before gravity and grid-stage collider/action updates is
\begin{equation}
\bm u_a=\bm p_a/m_a,\qquad m_a>0.
\label{eq:gridupdate}
\end{equation}
Here $w_{aq}$ is the interpolation weight, $\bm x_a$ the grid-node position, $\bm B_q$ the APIC affine velocity map, and $\bm f_a^{\mathrm{int}}$ the stress force. Gravity and collider/action updates precede G2P. Every particle samples the common field, providing grid-mediated inter-object coupling; configured collider friction handles collider interactions. $\Gamma^{(c)}$ schedules substep actions and output-frame projections.

\paragraph{Kinematic structure-preserving projection for support-like parts.}
For a row $r$ selected as support-like by the fixed structural-role predicate, let $\particleset_r=\{q\mid z_q^Q=r\}$ and $\bm\delta_{qr}^0=\bm x_q^0-\bar{\bm x}_r^0$. At rendered-frame solver step $n_t$, $\bm x_q$, $\bm v_q$, $\bm F_q$, and $\bm B_q$ denote position, velocity, deformation gradient, and affine velocity matrix; tildes are provisional states and bars are means over $\particleset_r$. The relaxed projection is
\begin{equation}
\begin{aligned}
\bm x_q^{n_t}
&=(1-\eta_r^{(c)})\widetilde{\bm x}_q^{n_t}
+\eta_r^{(c)}(\widetilde{\bar{\bm x}}_r^{n_t}+\bm\delta_{qr}^0),\\
\bm v_q^{n_t}
&=(1-\eta_r^{(c)})\widetilde{\bm v}_q^{n_t}
+\eta_r^{(c)}\widetilde{\bar{\bm v}}_r^{n_t}.
\end{aligned}
\label{eq:structuremotion}
\end{equation}
The same strength relaxes the deformation gradient and affine velocity state:
\begin{equation}
\begin{aligned}
\bm F_q^{n_t}
&=(1-\eta_r^{(c)})\widetilde{\bm F}_q^{n_t}+\eta_r^{(c)}\bm I_3,\\
\bm B_q^{n_t}
&=(1-\eta_r^{(c)})\widetilde{\bm B}_q^{n_t}.
\end{aligned}
\label{eq:structurestate}
\end{equation}
With relaxation weight $\eta_r^{(c)}\in[0,1]$ supplied by $\Gamma^{(c)}$, this optional operator regularizes the provisional state toward the compiled rest-shape prior, reducing local stretch/shear associated with burr-like protrusions after Gaussian transport \citep{xie2024physgaussian,feng2025gaussiansplashing}.

An optional interface-retention projection reduces visible separation between attached parts using a fixed frame-0 ``initial neighborhood.'' It and Equations~\eqref{eq:structuremotion}--\eqref{eq:structurestate} are applied after G2P at scheduled output frames and before Gaussian transport; the corrected state seeds the next interval. Attached-part cases may activate it, whereas multi-object cases use shared-grid interaction; details are in the supplement.

\paragraph{Transport to dynamic Gaussians.}
For the Gaussian-associated particle $i$, let $\bm F_{i,\mathrm{sc}}^t=\bm L^{-1}\bm F_i^{n_t}\bm L$. Since $\bm L=s\bm I_3$, this equals $\bm F_i^{n_t}$; the conjugate form records the coordinate conversion. Rendering uses
\begin{align}
\bm\mu_i^t&=\bm L^{-1}(\bm x_i^{n_t}-\bm d),\notag\\
\bm\Sigma_i^t&=\bm F_{i,\mathrm{sc}}^t\bm\Sigma_i^0
(\bm F_{i,\mathrm{sc}}^t)^{\mathsf T}.
\label{eq:gaussiantransport}
\end{align}
Opacity and appearance stay fixed; the polar rotation of $\bm F_{i,\mathrm{sc}}^t$ updates spherical-harmonic viewing direction \citep{xie2024physgaussian}. Gaussian-associated particles are rendered, while filled particles support simulation.

\begin{figure*}[!t]
\centering
\includegraphics[width=0.64\textwidth]{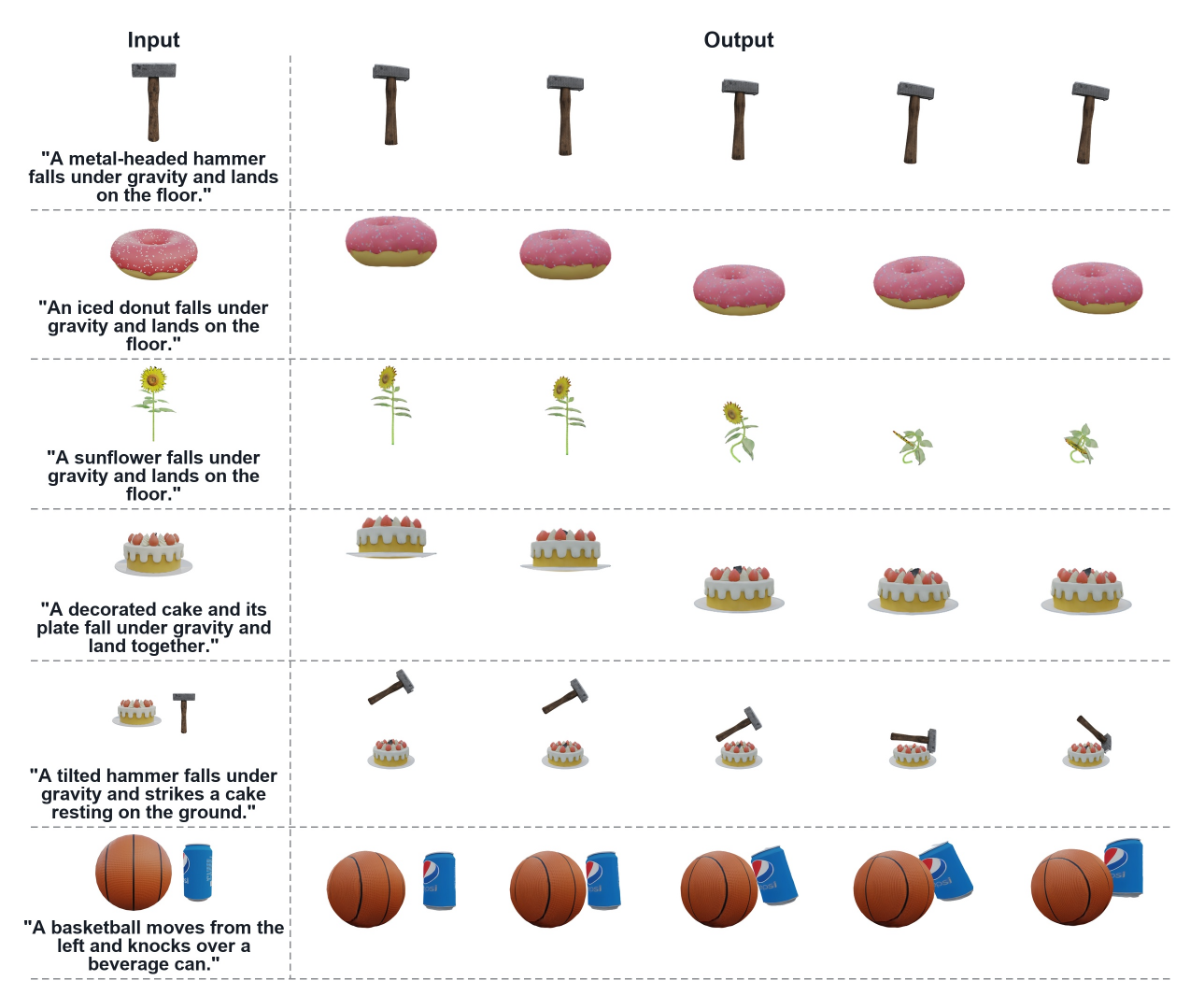}
\caption{PhysMAS rollouts with an input, prompt, and five synchronized frames per row. Rows 1--4 highlight part-dependent deformation; Rows 5--6 show multi-object execution.}
\label{fig:gallery}
\end{figure*}

\subsection{Executable Candidate Selection}
\label{sec:executioncheck}

For candidate $c$, deterministic skills return process, numerical, artifact, and endpoint diagnostics $\mathcal D^{(c)}$ together with rollout $\mathcal G^{(c,t)}$ and video $\mathcal V^{(c)}$. Algorithm~\ref{alg:selection} defines the bounded selection loop; Figure~\ref{fig:overview}'s feedback path routes diagnostics to fixed repairs while preserving row-wise material evidence.

\begin{algorithm}[!b]
\caption{PhysMAS inference and candidate selection}
\label{alg:selection}
\footnotesize
\begin{algorithmic}[1]
\Require prompt $y$, views $\mathcal I$, aligned $\mathcal G^0$, output cameras
\State $(\bm z^G,\Pi)\gets\mathcal A_{\mathrm{scene}}(y,\mathcal I,\mathcal G^0)$
\For{each $r\in\mathcal R^{\mathrm{act}}$}
  \State $\bm\psi_r\gets\mathcal A_{\mathrm{mat}}(\mathcal H_r)$
\EndFor
\State $\mathcal C\gets\operatorname{Build}(\{\bm\psi_r\},\Pi)$
\State $\mathcal E\gets\varnothing$
\For{each refinement round in the fixed finite schedule}
  \State execute every unseen $c\in\mathcal C$; add $(c,\mathcal D^{(c)})$ to $\mathcal E$
  \State $\mathcal A\gets\{c\in\mathcal E: \operatorname{Valid}(c)=1\land Q_{\mathrm{exec}}(c)\geq\tau\}$
  \If{$\mathcal A\neq\varnothing$}
    \State $\widehat c\gets\operatorname*{arg\,max}_{c\in\mathcal A}Q_{\mathrm{exec}}(c)$
    \State \Return $(\widehat c,\mathcal V^{(\widehat c)})$ if audited; otherwise failure
  \EndIf
  \If{the current round is not the final round}
    \State $b\gets\operatorname{RepairBase}(\mathcal E)$
    \State $\mathcal C\gets\operatorname{Unseen}(\operatorname{Repair}(b))$
  \EndIf
\EndFor
\State \Return failure
\end{algorithmic}
\end{algorithm}

Only the Object-Part Scene Agent may update $\bm z^G$ and $\Pi$, invoke tools, or accept a rollout; the Material Reasoning Agent returns $\bm\psi_r$, while compilation, simulation, repairs, gating, and ranking are fixed. The loop uses fixed candidate and repair budgets. $\operatorname{Valid}$ requires resolved actions, successful termination, stable numerical/runtime status, and fixed support, spread, and flattening criteria. $Q_{\mathrm{exec}}$ combines artifact availability, confidence, material/role compatibility, numerical-range terms, and optional-projection accounting, with a fixed acceptance threshold $\tau$. Both operations use execution records and compiled candidate metadata. Final audit confirms a decodable sequence and the multi-object interface-retention policy, yielding a reproducible executability criterion. Exact budgets, threshold, weights, criteria, repair order, and ties are in the supplement; representative rollouts appear in Figure~\ref{fig:gallery}.

\section{Experiments}
\label{sec:experiments}

\begin{table*}[!t]
\centering
{
\small
\setlength{\tabcolsep}{1.8pt}
\renewcommand{\arraystretch}{1.06}
\begin{tabular}{@{}l*{14}{c}@{}}
\toprule
\multirow{2}{*}{Method}
& \multicolumn{2}{c}{Jelly}
& \multicolumn{2}{c}{Metal}
& \multicolumn{2}{c}{Sand}
& \multicolumn{2}{c}{Foam}
& \multicolumn{2}{c}{Snow}
& \multicolumn{2}{c}{Plasticine}
& \multicolumn{2}{c}{Avg.} \\
\cmidrule(lr){2-3}\cmidrule(lr){4-5}\cmidrule(lr){6-7}\cmidrule(lr){8-9}
\cmidrule(lr){10-11}\cmidrule(lr){12-13}\cmidrule(lr){14-15}
& CLIP$_{sim}$ & UPR
& CLIP$_{sim}$ & UPR
& CLIP$_{sim}$ & UPR
& CLIP$_{sim}$ & UPR
& CLIP$_{sim}$ & UPR
& CLIP$_{sim}$ & UPR
& CLIP$_{sim}$ & UPR \\
\midrule
OmniPhysGS
& 0.2292 & 8\%
& 0.2151 & 6\%
& 0.2048 & 9\%
& 0.2053 & 7\%
& 0.1831 & 11\%
& 0.2135 & 7\%
& 0.2085 & 8.0\% \\
PhysSplat
& 0.2528 & 17\%
& 0.2426 & 15\%
& 0.2419 & 13\%
& 0.2390 & 16\%
& 0.2218 & 17\%
& 0.2516 & 19\%
& 0.2416 & 16.2\% \\
DreamPhysics
& 0.2459 & 12\%
& 0.2272 & 10\%
& 0.2216 & 8\%
& 0.2326 & 10\%
& 0.2071 & 13\%
& 0.2437 & 15\%
& 0.2297 & 11.3\% \\
PhysGM
& 0.2774 & 28\%
& 0.2732 & 29\%
& 0.2997 & 34\%
& 0.2609 & 23\%
& 0.2548 & 27\%
& 0.2691 & 25\%
& 0.2725 & 27.7\% \\
\midrule
\textbf{\ours}
& \textbf{0.2819} & \textbf{35\%}
& \textbf{0.2770} & \textbf{40\%}
& \textbf{0.3021} & \textbf{36\%}
& \textbf{0.2784} & \textbf{44\%}
& \textbf{0.2595} & \textbf{32\%}
& \textbf{0.2734} & \textbf{34\%}
& \textbf{0.2787} & \textbf{36.8\%} \\
\bottomrule
\end{tabular}
}
\caption{Multi-part results across six material profiles. Entries are CLIP$_{sim}$/UPR; ``Avg.'' is the six-profile macro-average. Multi-profile cases enter each applicable column. Higher is better.}
\label{tab:mainquant}
\vspace{-7pt}
\end{table*}

\subsection{Implementation Details}
\label{sec:implementation}

We implement \method{} in PyTorch with Qwen3.7-Plus as backbone and pretrained Grounding DINO and SAM~2 for perception. Each case follows the Problem Setup; on one RTX~4090, MPM uses a $100^3$ grid and at most 200,000 particles to render 50 frames at $800\times800$. The fixed run specification supplies gravity, colliders, and numerical settings; the agent supplies roles, targets, and supported actions. Candidates share this action specification and are generated independently of the paired reference videos used for evaluation. Further settings are in the supplement.

\subsection{Datasets and Evaluation Protocol}
\label{sec:data}

We evaluate preconstructed static 3DGS scenes from \textbf{PhysGaussian} \citep{xie2024physgaussian}, \textbf{Objaverse} \citep{deitke2023objaverse}, and \textbf{PhysAssets} \citep{lv2025physgm}. The \textbf{multi-part evaluation} uses single objects with multiple semantic parts and covers the six simulator profiles in Table~\ref{tab:mainquant}. Profile labels serve exclusively as aggregation categories; a multi-profile case enters each applicable column, and ``Avg.'' is the unweighted mean of the six profile scores.

The \textbf{multi-object evaluation} uses scenes with at least two identifiable objects and a visible impact or support interaction. It evaluates shared-grid execution for these interactions, with results aggregated jointly across scenes.

\subsection{Comparison with State-of-the-Art Methods}

\paragraph{Baselines.}
We compare with OmniPhysGS \citep{lin2025omniphysgs}, PhysSplat \citep{zhao2025physsplat}, DreamPhysics \citep{huang2025dreamphysics}, and PhysGM \citep{lv2025physgm}. Each receives the same motion prompt, initial 3DGS, cameras, external action, ground, timestamps, duration, and resolution through its native interface. Every baseline retains its native input representation under this shared evaluation setup.

\paragraph{Perceptual metrics.}
Following prior evaluations \citep{lv2025physgm,lv2026physagent}, we report CLIP similarity (CLIP$_{sim}$) and User Preference Rate (UPR). Frozen CLIP ViT-B/32 compares each frame with the complete motion prompt \citep{radford2021clip}; scores are averaged by case and reported set under the fixed aggregation protocol. UPR is forced-choice preference for prompt-consistent, perceptually plausible motion from 20 CV/CG researchers; method names are hidden and positions randomized. Main results use 5AFC and ablations use a separate 3AFC, which are reported separately because their choice sets differ (full protocol in the supplement). Together, they assess prompt alignment and perceived motion quality.

\paragraph{Motion and deformation metrics.}
Against paired references reserved for evaluation, \emph{trajectory nRMSE} compares image-normalized CoTracker3 tracks \citep{karaev2025cotracker3}; \emph{area log-RMSE}, mask areas; and \emph{local-strain RMSE}, frame-0-neighborhood Green--Lagrange strains \citep{kim2026physgaia}. Table~\ref{tab:physicalmetrics} reports 100 times each raw RMSE, macro-averaged over multi-part profiles or multi-object cases. Together, these diagnostics provide complementary image-plane evidence of motion and deformation.

\begin{table}[!b]
\centering
{
\small
\setlength{\tabcolsep}{2.4mm}
\begin{tabular}{@{}lcc@{}}
\toprule
Method & CLIP$_{sim}$ $\uparrow$ & UPR (\%) $\uparrow$\\
\midrule
OmniPhysGS & 0.2391 & 8.7\%\\
PhysSplat & 0.2434 & 13.3\%\\
DreamPhysics & 0.2437 & 15.0\%\\
PhysGM & 0.2631 & 25.7\%\\
\midrule
\textbf{\ours} & \textbf{0.2733} & \textbf{37.3\%}\\
\bottomrule
\end{tabular}
}
\caption{Multi-object results; higher is better.}
\label{tab:multiobject}
\vspace{-7pt}
\end{table}

\paragraph{Multi-part results.}
\label{sec:mainresults}
Table~\ref{tab:mainquant} shows a clear progression among the baselines. OmniPhysGS obtains the lowest macro-average (0.2085/8.0\%), DreamPhysics reaches 0.2297/11.3\%, and PhysSplat improves this to 0.2416/16.2\%. PhysGM is the strongest baseline at 0.2725/27.7\%, whereas \method{} achieves 0.2787/36.8\%. The profile-wise results are consistent: \method{} gives the highest CLIP$_{sim}$ and UPR for all six profiles rather than relying on one favorable material category. The larger UPR differences on Metal and Foam show that the improvement is also reflected in human preference, even when the corresponding CLIP$_{sim}$ margins are modest.

\paragraph{Multi-object results.}
Table~\ref{tab:multiobject} evaluates the methods separately when multiple objects participate in one interaction. OmniPhysGS records 0.2391/8.7\%, while PhysSplat and DreamPhysics yield similar CLIP$_{sim}$ values of 0.2434 and 0.2437 with UPRs of 13.3\% and 15.0\%. PhysGM improves both measures to 0.2631/25.7\%. \method{} ranks first at 0.2733/37.3\%, showing that its advantage extends from heterogeneous parts within one object to scene-level interactions involving distinct object identities.

\paragraph{Motion, deformation, and efficiency.}
The two entries in each cell of Table~\ref{tab:physicalmetrics} correspond to multi-part and multi-object results. For multi-part motion, PhysGM is the strongest baseline on all three diagnostics, while \method{} further reduces trajectory, projected-area, and local-strain error to 0.626, 2.641, and 6.654. The multi-object ranking is less uniform: DreamPhysics gives the lowest baseline trajectory and area errors, whereas OmniPhysGS gives the lowest baseline local-strain error. \method{} remains lowest on all three at 1.085, 3.980, and 15.361, respectively. Table~\ref{tab:runtime} complements these quality measures: \method{} matches PhysSplat's rounded two-minute runtime, is shorter than OmniPhysGS and DreamPhysics, but slower than the sub-minute PhysGM.

\begin{table}[!h]
\centering
{
\small
\setlength{\tabcolsep}{5mm}
\begin{tabular}{@{}lc@{}}
\toprule
Method & Runtime $\downarrow$\\
\midrule
OmniPhysGS & $>12$ h\\
PhysSplat & $2$ min\\
DreamPhysics & $>0.5$ h\\
PhysGM & $<1$ min\\
\ours & $2$ min\\
\bottomrule
\end{tabular}
}
\caption{Post-reconstruction runtime per 50-frame sequence; lower is better.}
\label{tab:runtime}
\vspace{-7pt}
\end{table}

\begin{table}[!t]
\centering
{
\small
\setlength{\tabcolsep}{1.2pt}
\renewcommand{\arraystretch}{1.06}
\begin{tabular}{@{}lccc@{}}
\toprule
Method & \makecell{Trajectory\\nRMSE} & \makecell{Area Log\\RMSE} & \makecell{Local-strain\\RMSE}\\
\midrule
OmniPhysGS   & 1.594/3.395 & 8.318/9.704 & 12.871/43.060\\
PhysSplat    & 1.447/3.192 & 7.821/9.113 & 15.959/46.563\\
DreamPhysics & 1.477/3.115 & 7.021/7.928 & 14.612/58.422\\
PhysGM       & 1.028/3.387 & 3.428/9.693 & 11.993/55.959\\
\midrule
\textbf{PhysMAS} & \textbf{0.626/1.085} & \textbf{2.641/3.980} & \textbf{6.654/15.361}\\
\bottomrule
\end{tabular}
}
\caption{Image-plane motion and deformation errors for multi-part/multi-object results; lower is better.}
\label{tab:physicalmetrics}
\vspace{-7pt}
\end{table}

\begin{figure}[!h]
\centering
\includegraphics[width=\columnwidth]{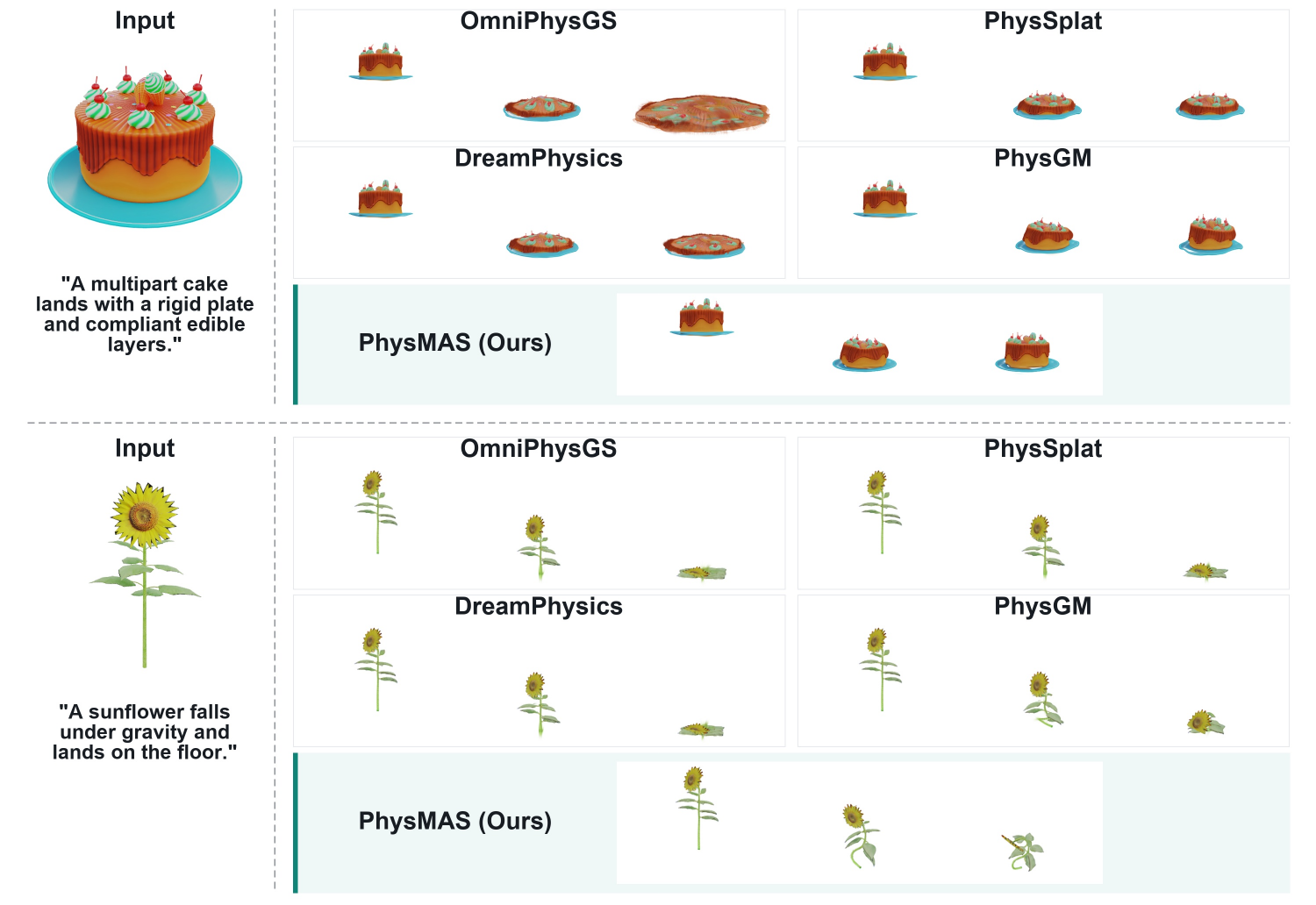}
\caption{Multi-part comparisons on cake impact and sunflower falling. Synchronized frames show part preservation across methods.}
\label{fig:qualitative}
\end{figure}

\paragraph{Qualitative results.}
Figure~\ref{fig:qualitative} supports the quantitative trends with synchronized multi-part examples. OmniPhysGS produces the strongest global flattening, PhysSplat retains more height but still exhibits largely uniform deformation, and DreamPhysics again converges to compressed endpoints. PhysGM better preserves the overall object silhouette, whereas \method{} more clearly retains the cake's plate/layer separation and the sunflower's part-dependent bending. The broader gallery in Figure~\ref{fig:gallery} separates the two settings: Rows~1--4 show material- and part-dependent responses within individual objects, while Rows~5--6 show the cake--hammer and basketball--beverage interactions with distinct objects maintained throughout the rollout.

\FloatBarrier

\subsection{Ablation Study}
\label{sec:ablations}

\begin{figure}[!h]
\centering
\includegraphics[width=\columnwidth]{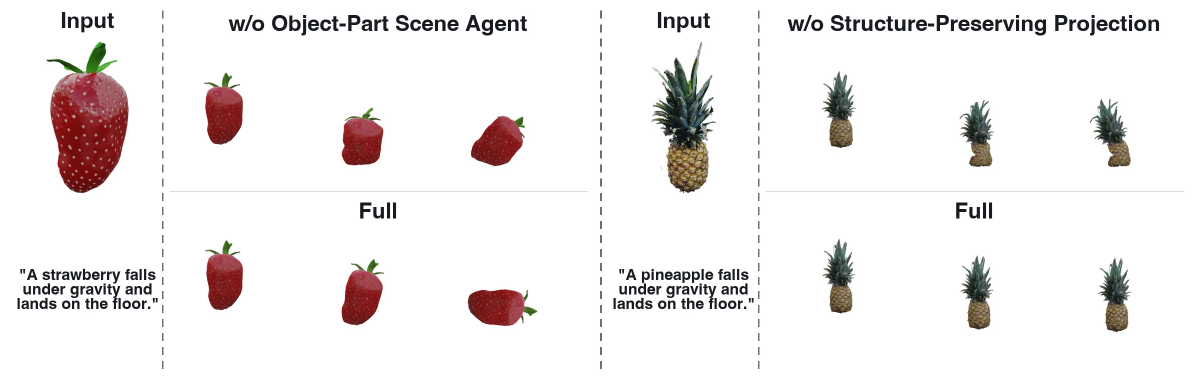}
\caption{Qualitative ablation for Table~\ref{tab:ablation}. Each case shows the input and synchronized frames from the indicated ablation (top) and Full PhysMAS (bottom).}
\label{fig:qualitative_ablation}
\end{figure}

\textbf{Object-Part Scene Agent.}
This bundle ablation replaces the agent-owned object--part state and part-aware compiler with an object-level variant. The full model is higher on both metrics for every profile, and macro CLIP$_{sim}$/UPR decreases from 0.2787/42.5\% to 0.2616/25.5\%. Foam shows the largest profile-level change, from 0.2784/49\% to 0.2184/14\%, where a single object-wide assignment removes the intended distinction between its constituent regions. The strawberry sequences in Figure~\ref{fig:qualitative_ablation} provide the corresponding visual comparison across landing and contact. Together, these results support the combined contribution of agent-owned object--part state and part-aware compilation.

\begin{table}[!t]
\centering
{
\small
\setlength{\tabcolsep}{1.5pt}
\renewcommand{\arraystretch}{1.0}
\begin{tabular}{@{}lcccccc@{}}
\toprule
\multirow{2}{*}{Profile}
& \multicolumn{2}{c}{w/o O-P Agent}
& \multicolumn{2}{c}{w/o S-P Projection}
& \multicolumn{2}{c}{\textbf{Full}} \\
\cmidrule(lr){2-3}\cmidrule(lr){4-5}\cmidrule(lr){6-7}
& CLIP$_{sim}$ & UPR & CLIP$_{sim}$ & UPR & CLIP$_{sim}$ & UPR \\
\midrule
Jelly      & 0.2727 & 31\% & 0.2696 & 26\% & \textbf{0.2819} & \textbf{43\%} \\
Metal      & 0.2693 & 27\% & 0.2703 & 31\% & \textbf{0.2770} & \textbf{42\%} \\
Sand       & 0.2921 & 27\% & 0.3015 & 35\% & \textbf{0.3021} & \textbf{38\%} \\
Foam       & 0.2184 & 14\% & 0.2684 & 37\% & \textbf{0.2784} & \textbf{49\%} \\
Snow       & 0.2484 & 26\% & 0.2487 & 30\% & \textbf{0.2595} & \textbf{44\%} \\
Plasticine & 0.2688 & 28\% & 0.2719 & 33\% & \textbf{0.2734} & \textbf{39\%} \\
\midrule
Avg.       & 0.2616 & 25.5\% & 0.2717 & 32.0\% & \textbf{0.2787} & \textbf{42.5\%} \\
\bottomrule
\end{tabular}
}
\caption{Six-profile ablation; UPR is in percent and ``Avg.'' is the macro-average.}
\label{tab:ablation}
\vspace{-7pt}
\end{table}

\textbf{Structure-Preserving Projection.}
Removing the operator package lowers macro CLIP$_{sim}$/UPR to 0.2717/32.0\%, with the full model remaining higher for all six profiles. The UPR differences are most pronounced on Jelly, Snow, and Foam, where the full model gains 17, 14, and 12 percentage points, respectively. In Figure~\ref{fig:qualitative_ablation}, the variant without projection shows greater distortion of the pineapple's crown--body support structure, whereas the full method maintains a more stable silhouette through landing. This isolates the projection package's contribution to visual support-shape stabilization and complements the object--part ablation above.

\section{Discussion}
\label{sec:discussion}

PhysMAS fills a key gap in physics-grounded 4D Gaussian synthesis by automating object--part scene organization and part-wise material assignment. Persistent object--part identities and material profiles provide a structured interface between multimodal reasoning and MPM simulation, while solver-aware skills compile and screen executable candidates. Our implementation uses a $100^3$ grid to balance fidelity and efficiency. Future work will combine adaptive simulation with learned execution feedback to support real-time, large-scale compositional dynamics.

\section{Conclusion}
\label{sec:conclusion}

We introduce PhysMAS, a physics-grounded multi-agent framework for compositional 4D Gaussian synthesis. Its Object-Part Scene Agent, Material Reasoning Agent, and solver-aware skills carry persistent identities and part-wise profiles into shared-domain MPM simulation, supporting multi-part and multi-object dynamics without per-scene SDS backpropagation. Experiments demonstrate superior semantic alignment, perceived physical plausibility, and reference-video motion and deformation consistency among the compared methods while maintaining a short runtime.

\bibliography{references}

@article{kerbl2023gaussian,
  title={{3D} Gaussian Splatting for Real-Time Radiance Field Rendering},
  author={Kerbl, Bernhard and Kopanas, Georgios and Leimkuehler, Thomas and Drettakis, George},
  journal={ACM Transactions on Graphics},
  volume={42},
  number={4},
  pages={139:1--139:14},
  year={2023},
  doi={10.1145/3592433}
}

@inproceedings{wu2024fourDGS,
  title={{4D} Gaussian Splatting for Real-Time Dynamic Scene Rendering},
  author={Wu, Guanjun and Yi, Taoran and Fang, Jiemin and Xie, Lingxi and Zhang, Xiaopeng and Wei, Wei and Liu, Wenyu and Tian, Qi and Wang, Xinggang},
  booktitle={Proceedings of the IEEE/CVF Conference on Computer Vision and Pattern Recognition},
  pages={20310--20320},
  year={2024},
  url={https://openaccess.thecvf.com/content/CVPR2024/html/Wu_4D_Gaussian_Splatting_for_Real-Time_Dynamic_Scene_Rendering_CVPR_2024_paper.html}
}

@inproceedings{poole2023dreamfusion,
  title={{DreamFusion}: Text-to-{3D} Using {2D} Diffusion},
  author={Poole, Ben and Jain, Ajay and Barron, Jonathan T. and Mildenhall, Ben},
  booktitle={The Eleventh International Conference on Learning Representations},
  year={2023},
  url={https://openreview.net/forum?id=FjNys5c7VyY}
}

@inproceedings{yang2024deformablegs,
  title={Deformable {3D} Gaussians for High-Fidelity Monocular Dynamic Scene Reconstruction},
  author={Yang, Ziyi and Gao, Xinyu and Zhou, Wen and Jiao, Shaohui and Zhang, Yuqing and Jin, Xiaogang},
  booktitle={Proceedings of the IEEE/CVF Conference on Computer Vision and Pattern Recognition},
  pages={20331--20341},
  year={2024},
  url={https://openaccess.thecvf.com/content/CVPR2024/html/Yang_Deformable_3D_Gaussians_for_High-Fidelity_Monocular_Dynamic_Scene_Reconstruction_CVPR_2024_paper.html}
}

@inproceedings{li2024spacetime,
  title={Spacetime Gaussian Feature Splatting for Real-Time Dynamic View Synthesis},
  author={Li, Zhan and Chen, Zhang and Li, Zhong and Xu, Yi},
  booktitle={Proceedings of the IEEE/CVF Conference on Computer Vision and Pattern Recognition},
  pages={8508--8520},
  year={2024},
  url={https://openaccess.thecvf.com/content/CVPR2024/html/Li_Spacetime_Gaussian_Feature_Splatting_for_Real-Time_Dynamic_View_Synthesis_CVPR_2024_paper.html}
}

@inproceedings{ling2024align,
  title={Align Your Gaussians: Text-to-{4D} with Dynamic {3D} Gaussians and Composed Diffusion Models},
  author={Ling, Huan and Kim, Seung Wook and Torralba, Antonio and Fidler, Sanja and Kreis, Karsten},
  booktitle={Proceedings of the IEEE/CVF Conference on Computer Vision and Pattern Recognition},
  pages={8576--8588},
  year={2024},
  url={https://openaccess.thecvf.com/content/CVPR2024/html/Ling_Align_Your_Gaussians_Text-to-4D_with_Dynamic_3D_Gaussians_and_Composed_CVPR_2024_paper.html}
}

@inproceedings{jiang2024animate3d,
  title={{Animate3D}: Animating Any {3D} Model with Multi-View Video Diffusion},
  author={Jiang, Yanqin and Yu, Chaohui and Cao, Chenjie and Wang, Fan and Hu, Weiming and Gao, Jin},
  booktitle={Advances in Neural Information Processing Systems},
  volume={37},
  pages={125879--125906},
  year={2024},
  url={https://proceedings.neurips.cc/paper_files/paper/2024/hash/e3b53f89136b1bc69a5714ea465f01b6-Abstract-Conference.html}
}

@inproceedings{zhang2024fourdiffusion,
  title={{4Diffusion}: Multi-View Video Diffusion Model for {4D} Generation},
  author={Zhang, Haiyu and Chen, Xinyuan and Wang, Yaohui and Liu, Xihui and Wang, Yunhong and Qiao, Yu},
  booktitle={Advances in Neural Information Processing Systems},
  volume={37},
  pages={15272--15295},
  year={2024},
  url={https://proceedings.neurips.cc/paper_files/paper/2024/hash/1bbfea488a8968e2d3c6565639b08e5e-Abstract-Conference.html}
}

@inproceedings{ren2024l4gm,
  title={{L4GM}: Large {4D} Gaussian Reconstruction Model},
  author={Ren, Jiawei and Xie, Kevin and Mirzaei, Ashkan and Liang, Hanxue and Zeng, Xiaohui and Kreis, Karsten and Liu, Ziwei and Torralba, Antonio and Fidler, Sanja and Kim, Seung Wook and Ling, Huan},
  booktitle={Advances in Neural Information Processing Systems},
  volume={37},
  pages={56828--56858},
  year={2024},
  url={https://proceedings.neurips.cc/paper_files/paper/2024/hash/6808f2c57d9564a2639a4710e3bbd9b9-Abstract-Conference.html}
}

@inproceedings{wu2025cat4d,
  title={{CAT4D}: Create Anything in {4D} with Multi-View Video Diffusion Models},
  author={Wu, Rundi and Gao, Ruiqi and Poole, Ben and Trevithick, Alex and Zheng, Changxi and Barron, Jonathan T. and Holynski, Aleksander},
  booktitle={Proceedings of the IEEE/CVF Conference on Computer Vision and Pattern Recognition},
  pages={26057--26068},
  year={2025},
  url={https://openaccess.thecvf.com/content/CVPR2025/html/Wu_CAT4D_Create_Anything_in_4D_with_Multi-View_Video_Diffusion_Models_CVPR_2025_paper.html}
}

@inproceedings{xie2024physgaussian,
  title={{PhysGaussian}: Physics-Integrated {3D} Gaussians for Generative Dynamics},
  author={Xie, Tianyi and Zong, Zeshun and Qiu, Yuxing and Li, Xuan and Feng, Yutao and Yang, Yin and Jiang, Chenfanfu},
  booktitle={Proceedings of the IEEE/CVF Conference on Computer Vision and Pattern Recognition},
  pages={4389--4398},
  year={2024},
  url={https://openaccess.thecvf.com/content/CVPR2024/html/Xie_PhysGaussian_Physics-Integrated_3D_Gaussians_for_Generative_Dynamics_CVPR_2024_paper.html}
}

@inproceedings{feng2025gaussiansplashing,
  title={Gaussian Splashing: Unified Particles for Versatile Motion Synthesis and Rendering},
  author={Feng, Yutao and Feng, Xiang and Shang, Yintong and Jiang, Ying and Yu, Chang and Zong, Zeshun and Shao, Tianjia and Wu, Hongzhi and Zhou, Kun and Jiang, Chenfanfu and Yang, Yin},
  booktitle={Proceedings of the IEEE/CVF Conference on Computer Vision and Pattern Recognition},
  pages={518--529},
  year={2025},
  url={https://openaccess.thecvf.com/content/CVPR2025/html/Feng_Gaussian_Splashing_Unified_Particles_for_Versatile_Motion_Synthesis_and_Rendering_CVPR_2025_paper.html}
}

@inproceedings{zhang2024physdreamer,
  title={{PhysDreamer}: Physics-Based Interaction with {3D} Objects via Video Generation},
  author={Zhang, Tianyuan and Yu, Hong-Xing and Wu, Rundi and Feng, Brandon Y. and Zheng, Changxi and Snavely, Noah and Wu, Jiajun and Freeman, William T.},
  booktitle={European Conference on Computer Vision},
  pages={388--406},
  year={2024},
  doi={10.1007/978-3-031-72627-9_22}
}

@article{huang2025dreamphysics,
  title={{DreamPhysics}: Learning Physics-Based {3D} Dynamics with Video Diffusion Priors},
  author={Huang, Tianyu and Zhang, Haoze and Zeng, Yihan and Zhang, Zhilu and Li, Hui and Zuo, Wangmeng and Lau, Rynson W. H.},
  journal={Proceedings of the AAAI Conference on Artificial Intelligence},
  volume={39},
  number={4},
  pages={3733--3741},
  year={2025},
  doi={10.1609/aaai.v39i4.32389}
}

@inproceedings{lin2025omniphysgs,
  title={{OmniPhysGS}: {3D} Constitutive Gaussians for General Physics-Based Dynamics Generation},
  author={Lin, Yuchen and Lin, Chenguo and Xu, Jianjin and Mu, Yadong},
  booktitle={The Thirteenth International Conference on Learning Representations},
  year={2025},
  url={https://openreview.net/forum?id=9HZtP6I5lv}
}

@inproceedings{zhao2025physsplat,
  title={{PhysSplat}: Efficient Physics Simulation for {3D} Scenes via {MLLM}-Guided Gaussian Splatting},
  author={Zhao, Haoyu and Wang, Hao and Zhao, Xingyue and Fei, Hao and Wang, Hongqiu and Long, Chengjiang and Zou, Hua},
  booktitle={Proceedings of the IEEE/CVF International Conference on Computer Vision},
  pages={5242--5252},
  year={2025},
  url={https://openaccess.thecvf.com/content/ICCV2025/html/Zhao_PhysSplat_Efficient_Physics_Simulation_for_3D_Scenes_via_MLLM-Guided_Gaussian_ICCV_2025_paper.html}
}

@inproceedings{lv2025physgm,
  title={{PhysGM}: Large Physical Gaussian Model for Feed-Forward {4D} Synthesis},
  author={Lv, Chunji and Chen, Zequn and Di, Donglin and Zhang, Weinan and Li, Hao and Wei, Chen and Lei, Yinjie and Li, Changsheng},
  booktitle={Proceedings of the IEEE/CVF Conference on Computer Vision and Pattern Recognition},
  pages={29855--29865},
  year={2026},
  url={https://openaccess.thecvf.com/content/CVPR2026/html/Lv_PhysGM_Large_Physical_Gaussian_Model_for_Feed-Forward_4D_Synthesis_CVPR_2026_paper.html}
}

@misc{lv2026physagent,
  title={{PhysAgent}: Automating Physics-Based {4D} Synthesis via Trajectory-Grounded Multi-Agent Feedback},
  author={Lv, Chunji and Ye, Jiaxi and Jiang, Yuchen and Lin, Rexar and Li, Changsheng},
  year={2026},
  archivePrefix={arXiv},
  eprint={2606.08688}
}

@inproceedings{cai2024gic,
  title={{GIC}: Gaussian-Informed Continuum for Physical Property Identification and Simulation},
  author={Cai, Junhao and Yang, Yuji and Yuan, Weihao and He, Yisheng and Dong, Zilong and Bo, Liefeng and Cheng, Hui and Chen, Qifeng},
  booktitle={Advances in Neural Information Processing Systems},
  volume={37},
  pages={75035--75063},
  year={2024},
  url={https://proceedings.neurips.cc/paper_files/paper/2024/hash/89379d5fc6eb34ff98488202fb52b9d0-Abstract-Conference.html}
}

@inproceedings{li2025freegave,
  title={{FreeGave}: {3D} Physics Learning from Dynamic Videos by Gaussian Velocity},
  author={Li, Jinxi and Song, Ziyang and Zhou, Siyuan and Yang, Bo},
  booktitle={Proceedings of the IEEE/CVF Conference on Computer Vision and Pattern Recognition},
  pages={12433--12443},
  year={2025},
  url={https://openaccess.thecvf.com/content/CVPR2025/html/Li_FreeGave_3D_Physics_Learning_from_Dynamic_Videos_by_Gaussian_Velocity_CVPR_2025_paper.html}
}

@inproceedings{chen2025vid2sim,
  title={{Vid2Sim}: Generalizable, Video-Based Reconstruction of Appearance, Geometry and Physics for Mesh-Free Simulation},
  author={Chen, Chuhao and Dou, Zhiyang and Wang, Chen and Huang, Yiming and Chen, Anjun and Feng, Qiao and Gu, Jiatao and Liu, Lingjie},
  booktitle={Proceedings of the IEEE/CVF Conference on Computer Vision and Pattern Recognition},
  pages={26545--26555},
  year={2025},
  url={https://openaccess.thecvf.com/content/CVPR2025/html/Chen_Vid2Sim_Generalizable_Video-based_Reconstruction_of_Appearance_Geometry_and_Physics_for_CVPR_2025_paper.html}
}

@inproceedings{jiang2025phystwin,
  title={{PhysTwin}: Physics-Informed Reconstruction and Simulation of Deformable Objects from Videos},
  author={Jiang, Hanxiao and Hsu, Hao-Yu and Zhang, Kaifeng and Yu, Hsin-Ni and Wang, Shenlong and Li, Yunzhu},
  booktitle={Proceedings of the IEEE/CVF International Conference on Computer Vision},
  pages={7219--7230},
  year={2025},
  url={https://openaccess.thecvf.com/content/ICCV2025/html/Jiang_PhysTwin_Physics-Informed_Reconstruction_and_Simulation_of_Deformable_Objects_from_Videos_ICCV_2025_paper.html}
}

@inproceedings{liu2024physgen,
  title={{PhysGen}: Rigid-Body Physics-Grounded Image-to-Video Generation},
  author={Liu, Shaowei and Ren, Zhongzheng and Gupta, Saurabh and Wang, Shenlong},
  booktitle={European Conference on Computer Vision},
  pages={360--378},
  year={2024}
}

@article{wang2026motionphysics,
  title={{MotionPhysics}: Learnable Motion Distillation for Text-Guided Simulation},
  author={Wang, Miaowei and Zadrozny, Jakub and Mac Aodha, Oisin and Vaxman, Amir},
  journal={Proceedings of the AAAI Conference on Artificial Intelligence},
  volume={40},
  number={12},
  pages={9993--10001},
  year={2026},
  doi={10.1609/aaai.v40i12.37965}
}

@inproceedings{li2023pacnerf,
  title={{PAC-NeRF}: Physics Augmented Continuum Neural Radiance Fields for Geometry-Agnostic System Identification},
  author={Li, Xuan and Qiao, Yi-Ling and Chen, Peter Yichen and Jatavallabhula, Krishna Murthy and Lin, Ming C. and Jiang, Chenfanfu and Gan, Chuang},
  booktitle={The Eleventh International Conference on Learning Representations},
  year={2023},
  url={https://openreview.net/forum?id=tVkrbkz42vc}
}

@inproceedings{ma2024eureka,
  title={{Eureka}: Human-Level Reward Design via Coding Large Language Models},
  author={Ma, Yecheng Jason and Liang, William and Wang, Guanzhi and Huang, De-An and Bastani, Osbert and Jayaraman, Dinesh and Zhu, Yuke and Fan, Linxi and Anandkumar, Anima},
  booktitle={The Twelfth International Conference on Learning Representations},
  year={2024},
  url={https://openreview.net/forum?id=IEduRUO55F}
}

@inproceedings{ye2024gaussiangrouping,
  title={Gaussian Grouping: Segment and Edit Anything in {3D} Scenes},
  author={Ye, Mingqiao and Danelljan, Martin and Yu, Fisher and Ke, Lei},
  booktitle={European Conference on Computer Vision},
  pages={162--179},
  year={2024},
  doi={10.1007/978-3-031-73397-0_10}
}

@article{cen2025saga,
  title={Segment Any {3D} Gaussians},
  author={Cen, Jiazhong and Fang, Jiemin and Yang, Chen and Xie, Lingxi and Zhang, Xiaopeng and Shen, Wei and Tian, Qi},
  journal={Proceedings of the AAAI Conference on Artificial Intelligence},
  volume={39},
  number={2},
  pages={1971--1979},
  year={2025},
  doi={10.1609/aaai.v39i2.32193}
}

@inproceedings{qin2024langsplat,
  title={{LangSplat}: {3D} Language Gaussian Splatting},
  author={Qin, Minghan and Li, Wanhua and Zhou, Jiawei and Wang, Haoqian and Pfister, Hanspeter},
  booktitle={Proceedings of the IEEE/CVF Conference on Computer Vision and Pattern Recognition},
  pages={20051--20060},
  year={2024}
}

@inproceedings{yu2026part2gs,
  title={{Part$^2$GS}: Part-Aware Modeling of Articulated Objects using {3D} Gaussian Splatting},
  author={Yu, Tianjiao and Shah, Vedant and Wahed, Muntasir and Shen, Ying and Nguyen, Kiet A. and Lourentzou, Ismini},
  booktitle={Proceedings of the IEEE/CVF Conference on Computer Vision and Pattern Recognition},
  pages={18913--18923},
  year={2026}
}

@inproceedings{qiu2024featuresplatting,
  title={Language-Driven Physics-Based Scene Synthesis and Editing via Feature Splatting},
  author={Qiu, Ri-Zhao and Yang, Ge and Zeng, Weijia and Wang, Xiaolong},
  booktitle={European Conference on Computer Vision},
  pages={368--383},
  year={2024},
  url={https://feature-splatting.github.io/}
}

@inproceedings{xu2025gaussianproperty,
  title={{GaussianProperty}: Integrating Physical Properties to {3D} Gaussians with {LMMs}},
  author={Xu, Xinli and Ge, Wenhang and Qiu, Dicong and Chen, ZhiFei and Yan, Dongyu and Liu, Zhuoyun and Zhao, Haoyu and Zhao, Hanfeng and Zhang, Shunsi and Liang, Junwei and Chen, Ying-Cong},
  booktitle={Proceedings of the IEEE/CVF International Conference on Computer Vision},
  pages={7231--7240},
  year={2025}
}

@inproceedings{chopra2026physgs,
  title={{PhysGS}: Bayesian-Inferred Gaussian Splatting for Physical Property Estimation},
  author={Chopra, Samarth and Liang, Jing and Seneviratne, Gershom and Manocha, Dinesh},
  booktitle={Proceedings of the IEEE/CVF Conference on Computer Vision and Pattern Recognition},
  pages={18980--18990},
  year={2026}
}

@inproceedings{huang2026gaussianfluent,
  title={{GaussianFluent}: Gaussian Simulation for Dynamic Scenes with Mixed Materials},
  author={Huang, Bei and Chen, Yixin and Lu, Ruijie and Zeng, Gang and Zha, Hongbin and Pei, Yuru and Huang, Siyuan},
  booktitle={Proceedings of the IEEE/CVF Conference on Computer Vision and Pattern Recognition},
  pages={21583--21593},
  year={2026}
}

@inproceedings{wang2025decoupledgaussian,
  title={{DecoupledGaussian}: Object-Scene Decoupling for Physics-Based Interaction},
  author={Wang, Miaowei and Zhang, Yibo and Xu, Weiwei and Ma, Rui and Zou, Changqing and Morris, Daniel},
  booktitle={Proceedings of the IEEE/CVF Conference on Computer Vision and Pattern Recognition},
  pages={11361--11372},
  year={2025}
}

@inproceedings{chen2025physgen3d,
  title={{PhysGen3D}: Crafting a Miniature Interactive World from a Single Image},
  author={Chen, Boyuan and Jiang, Hanxiao and Liu, Shaowei and Gupta, Saurabh and Li, Yunzhu and Zhao, Hao and Wang, Shenlong},
  booktitle={Proceedings of the IEEE/CVF Conference on Computer Vision and Pattern Recognition},
  pages={6178--6189},
  year={2025},
  url={https://openaccess.thecvf.com/content/CVPR2025/html/Chen_PhysGen3D_Crafting_a_Miniature_Interactive_World_from_a_Single_Image_CVPR_2025_paper.html}
}

@inproceedings{jiang2026physho,
  title={{PhysHO}: Physics-Based Dynamic {3D} Gaussian Human and Object from Monocular Video},
  author={Jiang, Suyi and Lee, Gim Hee},
  booktitle={Proceedings of the IEEE/CVF Conference on Computer Vision and Pattern Recognition},
  pages={32507--32517},
  year={2026},
  url={https://openaccess.thecvf.com/content/CVPR2026/html/Jiang_PhysHO_Physics-Based_Dynamic_3D_Gaussian_Human_and_Object_from_Monocular_CVPR_2026_paper.html}
}

@inproceedings{li2025trace,
  title={{TRACE}: Learning {3D} Gaussian Physical Dynamics from Multi-View Videos},
  author={Li, Jinxi and Song, Ziyang and Yang, Bo},
  booktitle={Proceedings of the IEEE/CVF International Conference on Computer Vision},
  pages={8820--8829},
  year={2025}
}

@inproceedings{liu2026mosiv,
  title={{MOSIV}: Multi-Object System Identification from Videos},
  author={Liu, Chunjiang and Wang, Xiaoyuan and Lin, Qingran and Xiao, Albert and Chen, Haoyu and Wen, Shizheng and Zhang, Hao and Qi, Lu and Yang, Ming-Hsuan and Jeni, Laszlo A. and Xu, Min and Zhao, Yizhou},
  booktitle={The Fourteenth International Conference on Learning Representations},
  year={2026},
  url={https://openreview.net/forum?id=0ylAe3Orfy}
}

@inproceedings{liu2024groundingdino,
  title={{Grounding DINO}: Marrying {DINO} with Grounded Pre-Training for Open-Set Object Detection},
  author={Liu, Shilong and Zeng, Zhaoyang and Ren, Tianhe and Li, Feng and Zhang, Hao and Yang, Jie and Jiang, Qing and Li, Chunyuan and Yang, Jianwei and Su, Hang and Zhu, Jun and Zhang, Lei},
  booktitle={European Conference on Computer Vision},
  year={2024}
}

@inproceedings{ravi2024sam2,
  title={{SAM 2}: Segment Anything in Images and Videos},
  author={Ravi, Nikhila and Gabeur, Valentin and Hu, Yuan-Ting and Hu, Ronghang and Ryali, Chaitanya and Ma, Tengyu and Khedr, Haitham and Raedle, Roman and Rolland, Chloe and Gustafson, Laura and Mintun, Eric and Pan, Junting and Alwala, Kalyan Vasudev and Carion, Nicolas and Wu, Chao-Yuan and Girshick, Ross and Dollar, Piotr and Feichtenhofer, Christoph},
  booktitle={The Thirteenth International Conference on Learning Representations},
  year={2025},
  url={https://proceedings.iclr.cc/paper_files/paper/2025/hash/45c1f6a8cbf2da59ebf2c802b4f742cd-Abstract-Conference.html}
}

@article{deitke2023objaverse,
  title={{Objaverse}: A Universe of Annotated {3D} Objects},
  author={Deitke, Matt and Schwenk, Dustin and Salvador, Jordi and Weihs, Luca and Michel, Oscar and VanderBilt, Eli and Schmidt, Ludwig and Ehsani, Kiana and Kembhavi, Aniruddha and Farhadi, Ali},
  journal={Proceedings of the IEEE/CVF Conference on Computer Vision and Pattern Recognition},
  pages={13142--13153},
  year={2023}
}

@inproceedings{radford2021clip,
  title={Learning Transferable Visual Models From Natural Language Supervision},
  author={Radford, Alec and Kim, Jong Wook and Hallacy, Chris and Ramesh, Aditya and Goh, Gabriel and Agarwal, Sandhini and Sastry, Girish and Askell, Amanda and Mishkin, Pamela and Clark, Jack and Krueger, Gretchen and Sutskever, Ilya},
  booktitle={International Conference on Machine Learning},
  pages={8748--8763},
  year={2021}
}

@inproceedings{kim2026physgaia,
  title={{PhysGaia}: A Physics-aware Benchmark with Multi-Body Interactions for Dynamic Novel View Synthesis},
  author={Kim, Mijeong and Kim, Gunhee and Choi, Jungyoon and Roh, Wonjae and Han, Bohyung},
  booktitle={Proceedings of the IEEE/CVF Conference on Computer Vision and Pattern Recognition},
  pages={22604--22614},
  year={2026}
}

@inproceedings{karaev2025cotracker3,
  title={{CoTracker3}: Simpler and Better Point Tracking by Pseudo-Labelling Real Videos},
  author={Karaev, Nikita and Makarov, Yuri and Wang, Jianyuan and Neverova, Natalia and Vedaldi, Andrea and Rupprecht, Christian},
  booktitle={Proceedings of the IEEE/CVF International Conference on Computer Vision},
  pages={6013--6022},
  year={2025}
}

@article{sulsky1994particle,
  title={A Particle Method for History-Dependent Materials},
  author={Sulsky, Deborah and Chen, Zhen and Schreyer, Howard L.},
  journal={Computer Methods in Applied Mechanics and Engineering},
  volume={118},
  number={1--2},
  pages={179--196},
  year={1994},
  doi={10.1016/0045-7825(94)90112-0}
}

@article{stomakhin2013snow,
  title={A Material Point Method for Snow Simulation},
  author={Stomakhin, Alexey and Schroeder, Craig and Chai, Lawrence and Teran, Joseph and Selle, Andrew},
  journal={ACM Transactions on Graphics},
  volume={32},
  number={4},
  pages={102:1--102:10},
  year={2013},
  doi={10.1145/2461912.2461948}
}

@article{jiang2015apic,
  title={The Affine Particle-in-Cell Method},
  author={Jiang, Chenfanfu and Schroeder, Craig and Selle, Andrew and Teran, Joseph and Stomakhin, Alexey},
  journal={ACM Transactions on Graphics},
  volume={34},
  number={4},
  article={51},
  pages={1--10},
  year={2015},
  doi={10.1145/2766996}
}

@inproceedings{liu2023partslip,
  title={{PartSLIP}: Low-Shot Part Segmentation for {3D} Point Clouds via Pretrained Image-Language Models},
  author={Liu, Minghua and Zhu, Yinhao and Cai, Hong and Han, Shizhong and Ling, Zhan and Porikli, Fatih and Su, Hao},
  booktitle={Proceedings of the IEEE/CVF Conference on Computer Vision and Pattern Recognition},
  pages={21736--21746},
  year={2023},
  url={https://openaccess.thecvf.com/content/CVPR2023/html/Liu_PartSLIP_Low-Shot_Part_Segmentation_for_3D_Point_Clouds_via_Pretrained_CVPR_2023_paper.html}
}

\appendix
\numberwithin{equation}{section}
\numberwithin{table}{section}

\section{Agent Contracts and Object--Part State}
\label{app:algorithm}

Algorithm~1 in the main paper gives the complete inference and selection flow. This section expands the two agent contracts and the deterministic object--part lifting stage. Only the Object-Part Scene Agent can update scene identity and plan $\Pi$, invoke tools, or accept a rollout. The Material Reasoning Agent returns row-wise evidence; candidate construction, compilation, simulation, repair generation, gating, ranking, and final audit are fixed routines.

\subsection{Generalized Object-Part Scene Agent Prompt}

The invariant instruction template below summarizes the prompt contract. Direction labels and calibrated cameras accompany the four-view sheet, and the aligned 3DGS is addressed through the persistent scene record.

\begin{quote}
\small
\textbf{Role:} scene-level object--part orchestrator.\\
\textbf{Input:} motion prompt, four direction-labeled scene views, camera records, and the registered scene state.\\
\textbf{Instructions:} identify objects before their parts; keep object IDs stable across views; create part IDs only within a parent object; preserve identically named parts from different objects as different rows; form a scene plan using only the supported action vocabulary; request material evidence independently for every active row; and accept a rollout only after deterministic execution screening.\\
\textbf{Output:} an object inventory, object-scoped part rows, and a proposed structured scene plan. Do not emit constitutive parameters or modify fixed numerical settings.
\end{quote}

The proposed scene plan records source and target identities, semantic roles, action type, direction, magnitude, and active interval. A deterministic compiler resolves the referenced rows, canonicalizes the action fields, and validates the plan before rollout. Gravity, colliders, cameras, grid resolution, and rendering settings remain in the fixed run specification rather than in the language-model output. A one-part object follows the same schema with one semantic row; a multi-object scene uses distinct object IDs while retaining one scene-level plan.

\subsection{Generalized Material Reasoning Agent Prompt}

The Material Reasoning Agent is called once per active object--part row. Its invariant instruction template is:

\begin{quote}
\small
\textbf{Role:} row-wise material-evidence reasoner.\\
\textbf{Input:} one persistent object--part key, its stored name and structural role, highlighted observations from the calibrated views, and the six accepted simulator profiles.\\
\textbf{Instructions:} assess visual-material evidence for the specified row; distinguish the visible material label from the simulator profile; return confidence-bearing evidence for the profile and continuous $E$/$\nu$ statistics; preserve the supplied identity; and abstain when a required semantic row has no valid evidence.\\
\textbf{Output:} evidence over visual labels and simulator profiles, together with continuous $E$ and $\nu$ statistics where available. Density is supplied by the selected visual-label prior during deterministic candidate construction. Do not change object/part identities, the scene plan, actions, cameras, or solver settings.
\end{quote}

The row evidence is serialized as $\bm\psi_r$ and validated before candidate construction. Table~\ref{tab:agent-contracts} lists the normalized records passed between the agents and fixed routines; it is the semantic schema used throughout the paper, independent of backend file names.

\begin{table*}[t]
\centering
{
\small
\begin{tabularx}{\textwidth}{@{}lXX@{}}
\toprule
Record & Required fields & Owner / consumer \\
\midrule
Object inventory & persistent object ID, name, scene role, view-linked queries & Scene Agent / perception and lifting \\
Object--part row & compound key $r=(o,p)$, part name, parent object, structural role, physics-group token, schema-expected visual labels, evidence references & Scene Agent / Material Agent and compiler \\
Scene plan $\Pi$ & action type, source/target keys, direction, magnitude, active interval & Scene Agent / action compiler \\
Material evidence $\bm\psi_r$ & visual-label evidence, six-profile evidence, $E$/$\nu$ statistics, confidence & Material Agent / candidate builder \\
Whole-object evidence & full-object visual label, mapped profile, $E$/$\nu$ statistics, profile-derived density prior & Fixed evidence routine / candidate builder \\
Candidate row $\Theta_r^{(c)}$ & profile branch $\beta_r^{(c)}$, $E_r^{(c)}$, $\nu_r^{(c)}$, $\rho_r^{(c)}$ & Fixed builder / particle compiler \\
Candidate controls $\mathcal U_{\mathrm{aux}}^{(c)}$ & support-projection weights, interface-retention flags & Fixed builder / schedule compiler \\
Candidate metadata & source-transform ID, parent candidate ID, fixed-order index & Fixed builder / repair and screen \\
Execution record $\mathcal D^{(c)}$ & process, numerical, artifact, and endpoint diagnostics & Simulator skills / deterministic screen and Scene Agent audit \\
\bottomrule
\end{tabularx}
}
\caption{Structured records and ownership in the PhysMAS inference protocol.}
\label{tab:agent-contracts}
\end{table*}

\subsection{Perception Configuration}

Perception allows at most two segmentation attempts. The initial schema query receives the front scene view; agent mode then uses one direction-labeled $2\times2$ grid of all four views for perception planning and repair. Material reasoning uses one highlighted multi-view sheet per row. Calls use deterministic inference, while pretrained Grounding DINO and SAM~2 provide object/part masks and confidence values under a fixed configuration. The front-view object inventory defines canonical query IDs. Projection-consistent Gaussian support and fixed query order provide deterministic matching for masks in the remaining views.

\subsection{Object--Part Lifting, Guards, and Ties}
\label{app:lifting}

The main paper writes object and part support as $e^{\mathrm{obj}}_{ov}$ and $e^{\mathrm{part}}_{rv}$. In the implementation, support is zero outside the selected mask and otherwise combines the stored mask confidence with normalized distance to the mask boundary. An area-normalized positive tie-break prevents exact zero scores. Multi-view-consistent neighbors regularize ambiguous assignments, followed by a residual-label cleanup pass. Every Gaussian retains either its selected semantic row or an object-scoped residual row with a persistent part ID.

We retain the main-paper notation $r=(o,p)$ for a compound object--part row, where $o$ and $p$ denote its object and part components. Let $\mathcal Q=\{1,\ldots,N_Q\}$ be the particle index set and $\mathcal Q_r=\{q\in\mathcal Q\mid z_q^Q=r\}$ the particles assigned to row $r$. The method input and final rollout remain scene-level as defined in the main paper. In Table~\ref{tab:lifting-settings}, $A_r$ is the accepted multi-view mask area for row $r$, and $n_r^{\mathrm{cand}}$ is the number of residual Gaussians eligible for its cleanup pass.

Table~\ref{tab:lifting-settings} lists the frozen implementation settings used across all reported data sources.

\begin{table}[t]
\centering
{
\small
\begin{tabularx}{\columnwidth}{@{}lX@{}}
\toprule
Setting & Frozen value \\
\midrule
Interior-distance support & range $[0.45,1.00]$ \\
Mask-confidence floor & $0.05$ \\
Residual confidence cap & $0.20$ \\
Area tie-break & $10^{-6}/\max(1,A_r)$ \\
Ambiguity / reliable margin & $0.18$ / $0.30$ \\
Neighborhood sizes & $12$ and $16$ \\
Residual rescue target & \makecell[l]{$\min\!\bigl(n_r^{\mathrm{cand}},$\\$\max(64,\operatorname{round}(0.003N_G))\bigr)$} \\
\bottomrule
\end{tabularx}
}
\caption{Frozen implementation settings for Gaussian object--part lifting.}
\label{tab:lifting-settings}
\end{table}

\section{Material Profiles and Particle Compilation}
\label{app:material}

\subsection{Visual-to-Solver Mapping and Parameter Domain}

Table~\ref{tab:mapping} gives the visual-material priors and initial map used during material reasoning and deterministic candidate construction. The serialized $E$ and profile-derived $\rho$ values carry the nominal Pa and $\mathrm{kg/m^3}$ labels expected by the backend. After uniform geometry normalization, these values serve as canonical solver coefficients. Ordinary part rows constrain $E$ and $\nu$ to the selected visual material's supported domain and use the local solver ranges $\nu\in[0.05,0.45]$ and $\rho\in[300,3000]$. The explicitly listed composite transforms in Table~\ref{tab:candidatetransforms} instead use the whole-object record, allowing $\nu$ up to $0.485$ and a profile-density prior in $[100,7800]$. Thus $\mathcal K_{\mathrm{sim}}$ in the main paper is the union of the ordinary-row domain and these enumerated composite-transform domains, rather than the result of one contradictory global clamp. The adaptive wave-speed rule uses a minimum substep of $5\times10^{-5}$ seconds and at most 800 substeps per output frame.

\begin{table*}[t]
\centering
{
\small
\setlength{\tabcolsep}{9pt}
\begin{tabular}{@{}lcccc@{}}
\toprule
Visual label & Solver branch & $E$ prior range (Pa) & $\nu$ bound & Raw $\rho$ \\
\midrule
Wood & Metal & $10^8$--$2\!\times\!10^{10}$ & 0.25--0.45 & 700 \\
Metal & Metal & $10^9$--$3\!\times\!10^{11}$ & 0.20--0.35 & 7800 \\
Plastic & Metal & $10^6$--$5\!\times\!10^9$ & 0.30--0.45 & 1200 \\
Glass & Metal & $10^9$--$10^{11}$ & 0.18--0.30 & 2500 \\
Fabric & Foam & $10^4$--$10^8$ & 0.20--0.45 & 500 \\
Leather & Foam & $10^5$--$10^9$ & 0.30--0.49 & 900 \\
Ceramic & Metal & $10^8$--$10^{11}$ & 0.15--0.35 & 2500 \\
Stone & Metal & $10^8$--$10^{11}$ & 0.10--0.35 & 2600 \\
Rubber & Jelly & $10^4$--$10^8$ & 0.40--0.499 & 1100 \\
Paper & Foam & $10^5$--$10^9$ & 0.20--0.45 & 800 \\
Sand & Sand & $10^3$--$10^7$ & 0.20--0.45 & 1600 \\
Snow & Snow & $10^3$--$10^7$ & 0.10--0.35 & 300 \\
Plasticine & Plasticine & $10^3$--$10^7$ & 0.25--0.49 & 2000 \\
Foam & Foam & $10^3$--$10^7$ & 0.10--0.45 & 100 \\
\bottomrule
\end{tabular}
}
\caption{Exact initial mapping, visual-material $E$ prior ranges, $\nu$ bounds, and density priors.}
\label{tab:mapping}
\end{table*}

\subsection{Six Solver Profiles}
For every particle, the compiler converts $(E,\nu)$ to Lam\'e parameters
\begin{equation}
\mu=\frac{E}{2(1+\nu)},\qquad
\lambda=\frac{E\nu}{(1+\nu)(1-2\nu)},
\label{eq:supp-lame}
\end{equation}
while $\rho$ determines particle mass through $M_q=\rho_qV_q^0$. Table~\ref{tab:constitutive} lists the stress and inelastic-update branches used by the six profiles. These names identify the simulator-facing constitutive branches associated with the inferred visual-material evidence.

\begin{table*}[t]
\centering
{
\small
\begin{tabularx}{\textwidth}{@{}lXX@{}}
\toprule
Profile & Elastic / trial stress & Inelastic deformation update \\
\midrule
Jelly & Compressible Neo-Hookean & None \\
Metal & Fixed-corotational elasticity & None \\
Sand & Hencky trial stress with Drucker--Prager projection ($k=0$, $\alpha=0.16$) & Sand return mapping \\
Foam & Hencky elasticity & StVK-based viscoplastic return mapping \\
Snow & Hencky trial stress with Drucker--Prager projection ($k=1000$, $\alpha=0.16$) & Stress-space projection \\
Plasticine & Hencky trial stress with Drucker--Prager projection ($k=5000$, $\alpha=0.16$) & Von-Mises-with-damage return mapping \\
\bottomrule
\end{tabularx}
}
\caption{Constitutive branches selected by the six simulator profiles. ``None'' means that the trial deformation gradient is retained before stress evaluation.}
\label{tab:constitutive}
\end{table*}

\subsection{Evidence Aggregation and Continuous Parameters}

For the evidence aggregation summarized by $\mathcal A_{\mathrm{mat}}$ in the main paper, the frozen source coefficients for schema, crop predictor, VLM, role, and whole-object evidence are $0.25$, $0.35$, $0.30$, $0.10$, and $0.05$, respectively. These are unnormalized evidence coefficients---their sum need not equal one---and the resulting label scores are normalized only after all sources are accumulated. Crop records are averaged within their source. Valid VLM records are confidence-weighted and divided by their count. Reported semantic rows use validated crop and predictor evidence before aggregation, while residual rows retain a unit Plastic fallback. The same fixed in-run evidence sources are used throughout the reported evaluation.

The role source scans the lowercase concatenation of the stored part name and physical role. The same vocabulary is used unchanged across all three reported data sources. Let $R_r(m)$ be the maximum confidence assigned to label $m$ by all matching rules, or zero if none matches. The fixed lookup is: head/blade/tip/edge/impact/metal $\mapsto$ Metal at $0.75$; handle/grip/shaft $\mapsto$ Wood at $0.55$ and Plastic at $0.45$; wheel/tire/tyre/sole $\mapsto$ Rubber at $0.80$; cushion/pad/pillow/foam $\mapsto$ Foam at $0.75$; cloth/fabric/lace/upper $\mapsto$ Fabric at $0.65$; plate/bowl/ceramic $\mapsto$ Ceramic at $0.55$ and Plastic at $0.35$; and frosting/cream/cake/soft $\mapsto$ Foam at $0.55$ and Plasticine at $0.35$. Multiple matching rules contribute evidence, while $R_r$ retains the maximum per label for the guard below.

Let $\mathcal M_{\mathrm{vis}}$ be the 14 visual labels in Table~\ref{tab:mapping}, let $\mathcal M_r^{\mathrm{exp}}$ be the deduplicated schema-expected labels stored with row $r$, and let $s_r(m)$ be the resulting nonnegative weighted evidence. If its sum is zero, the fixed fallback replaces the score vector by a unit score at Plastic. Define $Z_r=\sum_{m\in\mathcal M_{\mathrm{vis}}}s_r(m)$ after this fallback, so $Z_r>0$. The normalized posterior and selected label are
\begin{equation}
\begin{aligned}
P_r(m)&=\frac{s_r(m)}{Z_r},\\
\widetilde m_r&=\operatorname*{arg\,max}_{m\in\mathcal M_{\mathrm{vis}}}P_r(m).
\end{aligned}
\label{eq:supp-materialposterior}
\end{equation}
Thus $P_r\in\Delta^{13}$, and posterior ties follow the stored evidence-insertion order. The unconstrained mode is $m_r^{(0)}=\widetilde m_r$. When $\mathcal M_r^{\mathrm{exp}}$ is nonempty, let $m_r^{\mathrm{exp}}$ be its highest-posterior label, with the deduplicated expected-label order breaking ties, and define
\begin{align}
\operatorname{Cap}_{\mathrm{exp}}(r)
={}&\max\!\left\{1.75P_r(m_r^{\mathrm{exp}}),\right.\notag\\
&\left.\hspace{15mm}P_r(m_r^{\mathrm{exp}})+0.20\right\},\notag\\
\operatorname{ExpGuard}(r)
={}&[m_r^{(0)}\notin\mathcal M_r^{\mathrm{exp}}]
\land[P_r(m_r^{\mathrm{exp}})\geq0.18]\notag\\
&\land[P_r(m_r^{(0)})\leq
\operatorname{Cap}_{\mathrm{exp}}(r)],\notag\\
m_r^{(1)}={}&
\begin{cases}
m_r^{\mathrm{exp}},&\operatorname{ExpGuard}(r),\\
m_r^{(0)},&\text{otherwise}.
\end{cases}
\label{eq:supp-materialguard}
\end{align}
For an empty expected set, $\operatorname{ExpGuard}(r)$ is false.

Two guards then run in a fixed order. First, the multi-view VLM guard examines only records marked successful that do not abstain and provide normalized material probabilities, confidence at least $0.60$, and at least three canonical visible views. Let $v$ be a record's top label. The record is eligible only if its top-label probability is at least $0.75$, $v\in\mathcal M_r^{\mathrm{exp}}$, $v\neq m_r^{(1)}$, $R_r(v)\geq0.50$, $R_r(v)\geq R_r(m_r^{(1)})+0.10$, $P_r(v)\geq0.20$, $P_r(m_r^{(1)})\leq1.50P_r(v)$, and $P_r(m_r^{(1)})-P_r(v)\leq0.18$. If eligible records exist, $m_r^{(2)}$ is the label maximizing the product of VLM confidence, within-record top-label probability, and $P_r(v)$; otherwise $m_r^{(2)}=m_r^{(1)}$. Record order breaks ties.

Second, let $\operatorname{Map}$ denote the visual-label-to-branch map in Table~\ref{tab:mapping}, and let $\mathcal M_{\mathrm{el}}=\{m\mid\operatorname{Map}(m)=\mathrm{Metal}\}$. A row is structural when its normalized physics-group and role tokens contain support, supporting, structural, rigid, or loadbearing, or contain both load and bearing. A strong elastic-solid VLM record has at least three canonical visible views, confidence at least $0.60$, total probability on $\mathcal M_{\mathrm{el}}$ at least $0.70$, and maximum elastic-solid probability at least $0.50$. Define the expected-label elastic-solid fraction by
\begin{equation}
f_r^{\mathrm{el}}=
\begin{cases}
|\mathcal M_r^{\mathrm{exp}}\cap\mathcal M_{\mathrm{el}}|/
|\mathcal M_r^{\mathrm{exp}}|,&|\mathcal M_r^{\mathrm{exp}}|>0,\\
1,&|\mathcal M_r^{\mathrm{exp}}|=0.
\end{cases}
\label{eq:supp-expected-elastic}
\end{equation}
If $m_r^{(2)}\notin\mathcal M_{\mathrm{el}}$, the row is structural, $f_r^{\mathrm{el}}\geq0.50$, and such a record exists, then $m_r=\operatorname*{arg\,max}_{m\in\mathcal M_{\mathrm{el}}}P_r(m)$; otherwise $m_r=m_r^{(2)}$. Stored posterior order breaks the final tie. The stored material confidence is $P_r(m_r)$.

Residual rows with assigned Gaussians remain active during candidate construction. In the reported configuration, deserialization maps every particle to its stored active row, while the solver-compatible candidate uses its stored global profile. Ties follow active-row order.

For crop proposal $j$ of row $r$, the feed-forward model produces normalized means $\mu^E_{rj},\mu^\nu_{rj}$, uncertainty terms $\sigma^{E,\mathrm{model}}_{rj},\sigma^\nu_{rj}$, and visual-material evidence $e^{\mathrm{phys}}_{rj}$. The implementation decodes
\begin{align}
\widehat E_{rj}
&=0.1\,10^{2.456477\mu^E_{rj}+7.387210},\notag\\
\widehat\nu_{rj}
&=0.111\mu^\nu_{rj}+0.398.
\label{eq:decode}
\end{align}
The Qwen query provides discrete visual-material evidence, while continuous $E$ and $\nu$ statistics are pooled from validated crop predictions. Each crop contributes
$x^E_{rj}=\log_{10}\widehat E_{rj}$, $x^\nu_{rj}=\widehat\nu_{rj}$,
$w_{rj}=\max(0.05,e^{\mathrm{phys}}_{rj}(m_r))$, using $0.1$ when the selected-label entry is absent, and uncertainty
$a_{rj}=\sigma^{E,\mathrm{model}}_{rj}$. Let $\mathcal L_r^E$ and $\mathcal L_r^\nu$ contain the records finite for the $E$- and $\nu$-statistics, respectively, and define $W_r^E=\sum_{j\in\mathcal L_r^E}w_{rj}$ and $W_r^\nu=\sum_{j\in\mathcal L_r^\nu}w_{rj}$. For positive corresponding normalizers,
\begin{align}
\bar e_r
&=(W_r^E)^{-1}\sum_{j\in\mathcal L_r^E}w_{rj}x^E_{rj},\notag\\
\bar\nu_r
&=(W_r^\nu)^{-1}\sum_{j\in\mathcal L_r^\nu}w_{rj}x^\nu_{rj},\notag\\
v_r^e
&=(W_r^E)^{-1}\sum_{j\in\mathcal L_r^E}
w_{rj}(x^E_{rj}-\bar e_r)^2,\notag\\
s_r^e&=\max\!\left(0.25,\sqrt{v_r^e}\right).
\label{eq:moments}
\end{align}
Let $E_{\mathrm{def}}(m)$, $\nu_{\mathrm{def}}(m)$, and $\rho_{\mathrm{def}}(m)$ denote the geometric-midpoint $E$, midpoint $\nu$, and raw density for visual label $m$ in Table~\ref{tab:mapping}. With fewer than two valid $E$ records, $s_r^e=0.25$; with no valid $E$ record, the implementation sets $\bar e_r=\log_{10}E_{\mathrm{def}}(m_r)$ without evaluating the $E$ moments in Equation~\eqref{eq:moments}. With no valid $\nu$ record, it independently sets $\bar\nu_r=\nu_{\mathrm{def}}(m_r)$. Let $\operatorname{clip}^{E}_{m}$ and $\operatorname{clip}^{\nu}_{m}$ denote the $E$- and $\nu$-domain clamps of visual label $m$. The posterior builder then stores
\begin{align}
e_r^{\mathrm{post}}
&=\log_{10}\!\left[
\operatorname{clip}^{E}_{m_r}(10^{\bar e_r})
\right],\notag\\
\nu_r^{\mathrm{post}}
&=\operatorname{clip}^{\nu}_{m_r}(\bar\nu_r).
\label{eq:supp-posteriorcenter}
\end{align}
Let $\bar a_r=|\mathcal L_r^E|^{-1}\sum_{j\in\mathcal L_r^E}a_{rj}$ when $E$ records exist (default $0.35$ otherwise), $H_r=-\sum_mP_r(m)\log P_r(m)$ with $0\log0:=0$, and $h_r=\min(0.5,H_r/\max(1,\log14))$. Let $\bar\sigma_r^\nu$ be the mean finite crop uncertainty over $\mathcal L_r^\nu$, with default $0.06$ for an empty set. These defaults provide complete residual-row records. The final dispersions are
\begin{align}
\sigma_r^e
&=\max\!\left(
0.10,0.50\bar a_r+0.35s_r^e+0.15h_r\right),\notag\\
\sigma_r^\nu
&=\max\!\left(0.02,\bar\sigma_r^\nu\right).
\label{eq:dispersion}
\end{align}
Let $\gamma_r^{\mathrm{mask}}\in[0,1]$ be the stored part-mask confidence. The lifting stage applies the positive floor in Table~\ref{tab:lifting-settings}; consequently, a nonpositive serialized value is the sentinel for an unavailable confidence rather than an observed zero-confidence mask. Define
\begin{equation}
\widetilde\gamma_r^{\mathrm{mask}}=
\begin{cases}
\gamma_r^{\mathrm{mask}},&\gamma_r^{\mathrm{mask}}>0,\\
0.7,&\text{otherwise}.
\end{cases}
\label{eq:supp-maskconfidence}
\end{equation}
The stored posterior confidence used by candidate construction is
\begin{equation}
\gamma_r^{\mathrm{part}}=
\operatorname{clip}_{[0,1]}\!\left(
\frac{P_r(m_r)}{1+\sigma_r^e}
\max(0.2,\widetilde\gamma_r^{\mathrm{mask}})
\right).
\label{eq:supp-partconfidence}
\end{equation}
Here $\operatorname{clip}_{[0,1]}(x)=\min(1,\max(0,x))$.
For independent $\xi_E,\xi_\nu\in\{0.15,0.25,0.50,0.75,0.85\}$, the implementation uses the corresponding $\kappa_\xi\in\{-1.04,-0.67,0,0.67,1.04\}$ and computes
\begin{align}
E_r(\xi_E)&=\operatorname{clip}^{E}_{m_r}\!\left(
10^{e_r^{\mathrm{post}}+\kappa_{\xi_E}\sigma_r^e}\right),\notag\\
\nu_r(\xi_\nu)&=\operatorname{clip}^{\nu}_{m_r}
\!\left(\nu_r^{\mathrm{post}}+\kappa_{\xi_\nu}\sigma_r^\nu\right).
\label{eq:quantile}
\end{align}
A changed visual label uses its corresponding defaults before the same constraints are applied.

\subsection{Candidate Construction and Repair Transforms}
The initial row is
$(\beta_r^0,E_r^0,\nu_r^0,\rho_r^0)
=(\operatorname{Map}(m_r),E_r(0.50),\nu_r(0.50),
\operatorname{clip}(\rho_{\mathrm{def}}(m_r);300,3000))$.
Thus $\rho_r^0$ comes from the selected visual label's density prior rather than from a language-model prediction. The auxiliary whole-object record $(\beta_g,E_g,\nu_g,\rho_g)$ is produced from the full-object crop by the same visual mapping and crop-conditioned $E$/$\nu$ predictor, with density again supplied by its selected visual-label prior and all fields passed through the backend guard. When that record is absent, the fixed fallback is $(\mathrm{Plasticine},5\!\times\!10^5,0.476,5000)$. All predicate tests below are case-insensitive substring matches. Let $S_r$ denote the candidate-construction support predicate. A support token in the concatenated name, physics-group token, and role makes $S_r=1$. Only when none matches, a soft token makes $S_r=0$; if neither set matches, Wood, Metal, Glass, Ceramic, or Stone visual evidence makes $S_r=1$; it is zero otherwise. The support tokens are plate, dish, tray, stand, support, holder, structural, rigid, load bearing, load-bearing, frame, shell, blade, head, tip, and impact part. The soft tokens are soft, deformable, compressible, cushion, foam, sponge, flesh, fruit, food, filling, cream, frosting, and icing.

The candidate-construction cohesion predicate $C_r$ is evaluated only when $S_r=0$. It is one when the concatenated name, physics-group token, role, and visual label contains body, base, core, main, bulk, filling, coating, layer, cream, frosting, icing, fruit, organic, soft, deformable, foam, sponge, rubber, gel, cloth, fabric, leather, or pad. Otherwise it is one when $\beta_r^0\in\{\mathrm{Foam},\mathrm{Plasticine},\mathrm{Jelly}\}$ and the visual label is not Wood, Metal, Glass, Ceramic, or Stone; it is zero in every remaining case, including $S_r=1$. No other Sand or Snow exclusion is applied during candidate construction. The first matching rule defines $\omega_r$: it is 1.6 when $S_r=1$, 1.45 for candle/stem/stick/handle, 1.25 for fruit/strawberry/berry/decoration, 1.05 for cream/frosting/icing, and 1 otherwise.

Let $m_r^{\mathrm{sec}}$ be the second-ranked label in the stored posterior order and $E_r^{\mathrm{post}}=10^{e_r^{\mathrm{post}}}$. A row is alternative-feasible when $P_r(m_r^{\mathrm{sec}})\geq0.12$ and $E_{\mathrm{def}}(m_r^{\mathrm{sec}})\leq\max(10E_r^{\mathrm{post}},10^7)$. Among feasible rows, $r^\dagger$ is the row with the largest $P_r(m_r^{\mathrm{sec}})$; ties use the fixed active-row order, and no alternative candidate is emitted when this set is empty. All rows begin from their posterior medians. Row $r^\dagger$ receives $m_r^{\mathrm{sec}}$; if this differs from the stored label $m_r$, its mapped branch and default $E$, $\nu$, and density are used, whereas equality retains the posterior median. The ordinary physical and solver clamps follow in either case. The uncertainty sweep independently chooses the minimum-$\gamma_r^{\mathrm{part}}$ row with the fixed order as a tie-break.

In Table~\ref{tab:candidatetransforms}, $\operatorname{Map}$ and the three $\mathrm{def}$ functions use the branch and defaults just defined. Define $\beta_g^{\mathrm{coh}}=\beta_g$ when $\beta_g\in\{\mathrm{Plasticine},\mathrm{Jelly},\mathrm{Foam}\}$, and $\beta_g^{\mathrm{coh}}=\mathrm{Plasticine}$ otherwise. Initial-candidate rows use $S_r$ and $C_r$. A repair clones one rejected base candidate; let $(\beta_r^{(b)},E_r^{(b)},\nu_r^{(b)},\rho_r^{(b)})$ denote that clone's current row before the repair. Repair rows use the evaluator predicates $\widehat S_r(k)$ and $\widehat C_r(k)$ defined below. Table~\ref{tab:candidatetransforms} records the transforms; fields not named by a transform retain their current initial- or base-candidate values. Ordinary rows retain the local ranges above, while hard-support-bonded and solver-compatible rows use the explicitly shown whole-object $\nu$ and $\rho$ fields.

\begin{table*}[!t]
\centering
{
\small
\begin{tabularx}{\textwidth}{@{}p{0.18\textwidth}p{0.18\textwidth}X@{}}
\toprule
Candidate & Eligible row & Final solver row/action \\
\midrule
Posterior & all & $\beta=\beta^0$; $(\xi_E,\xi_\nu)=(0.50,0.50)$ \\
Soft / stiff & all & $\beta=\beta^0$; $(\xi_E,\xi_\nu)=(0.25,0.75)$ / $(0.75,0.50)$ \\
Uncertainty sweep & all & selected weakest-confidence row uses $(0.15,0.75)$; others use $(0.50,0.75)$ \\
Alternative material & $r=r^\dagger$ & \makecell[l]{use $m_r^{\mathrm{sec}}$; use its defaults only when $m_r^{\mathrm{sec}}\neq m_r$;\\apply the ordinary physical and solver clamps} \\
Target-impact & target row with $S_r=1$ & \makecell[l]{start from $(\xi_E,\xi_\nu)=(0.50,0.50)$; $\beta=\mathrm{Metal}$, $E=10^7$;\\$\nu=\operatorname{clip}(\nu^0;0.05,0.35)$, $\rho=\max(\rho^0,1000)$;\\disable projection and bonding} \\
Target-impact & target row with $S_r=0,C_r=1$ & \makecell[l]{start from $(0.50,0.50)$;\\$E=\operatorname{clip}(E^0;2\!\times\!10^4,7.5\!\times\!10^4)$;\\$\beta=\beta^0$ if $\beta^0\in\{\mathrm{Foam,Plasticine,Jelly}\}$, else Foam;\\$\nu=\operatorname{clip}(\nu^0;0.2,0.4)$; $\rho=\min(450,\max(300,\rho^0))$} \\
Target-impact & target row with $S_r=C_r=0$ & retain the posterior-median row and disable projection and bonding \\
Support-stiff & $S_r=1$ & \makecell[l]{$\beta=\mathrm{Metal}$, $E=10^7$; $\nu=\min(\nu^0,0.35)$,\\$\rho=\max(\rho^0,1000)$; no projection} \\
Hard-support bonded & $S_r=1$ & $\beta=\mathrm{Metal}$, $(E,\nu,\rho)=(10^7,0.35,2500)$; $\eta_r=0.85$ and enable interface retention \\
Hard-support bonded & $S_r=0$ & \makecell[l]{$\beta=\mathrm{Plasticine}$, $E=\max(E^0,\omega_rE_g)$;\\$\nu=\operatorname{clip}(\nu_g;0.2,0.485)$, $\rho=\rho_g$} \\
Solver-compatible & all & \makecell[l]{$\beta=\beta_g^{\mathrm{coh}}$; $E=\operatorname{clip}(\omega_rE_g;0.85E_g,1.7E_g)$;\\$\nu=\operatorname{clip}(\nu_g;0.2,0.485)$, $\rho=\rho_g$} \\
Rigid-support & all & \makecell[l]{start every row from $(\xi_E,\xi_\nu)=(0.25,0.75)$;\\for $S_r=1$, retain $\beta$ and current $\rho$, set $E=10^7$, $\eta_r=1$} \\
Repair: support stiffness & $\widehat S_r=1$ & $\beta=\mathrm{Metal}$, $E=10^7$, $\nu=\min(\nu^{(b)},0.35)$, $\rho=\max(\rho^{(b)},1000)$ \\
Repair: rigid support & $\widehat S_r=1$ & support-stiffness repair plus $\eta_r=1$ \\
Repair: elastic cohesion & $\widehat S_r=1$ & apply the support-stiffness repair \\
Repair: elastic cohesion & $\widehat S_r=0,\widehat C_r=1$ & \makecell[l]{$\beta=\mathrm{Jelly}$, $E=\max(E^{(b)},5\!\times\!10^5)$;\\$\nu=\operatorname{clip}(\nu^{(b)};0.2,0.42)$, $\rho=\max(\rho^{(b)},500)$} \\
Repair: higher cohesion & $\widehat S_r=1$ & apply the support-stiffness repair \\
Repair: higher cohesion & $\widehat S_r=0,\widehat C_r=1$ & retain $\beta^{(b)},\nu^{(b)}$; $E=\max(E^{(b)},5\!\times\!10^5)$, $\rho=\max(\rho^{(b)},500)$ \\
Repair: solver stability & all & retain $\beta^{(b)},E^{(b)}$; $\nu=\min(\nu^{(b)},0.40)$, $\rho=\max(\rho^{(b)},300)$ \\
Repair: role balanced & role-dependent & \makecell[l]{$\widehat S_r=1$ uses the support repair; $\widehat S_r=0,\widehat C_r=1$ with\\$\beta^{(b)}\in\{\mathrm{Foam,Plasticine}\}$ uses elastic-cohesion repair; others unchanged} \\
\bottomrule
\end{tabularx}
}
\caption{Implemented material-candidate transforms. $\operatorname{clip}(x;l,u)=\min(\max(x,l),u)$.}
\label{tab:candidatetransforms}
\end{table*}
The target-impact candidate is emitted only when $\Pi$ marks the interaction role as target, the supported action type specifies an impact, and at least one non-residual target row satisfies $S_r=0$ and $C_r=1$. Residual rows start from the same candidate quantile as other active rows but are skipped by the target-impact softening rule. The solver-compatible and hard-support-bonded candidates require at least one row with $S_r=1$ and another with $S_r=0,C_r=1$. Solver-compatible assigns the uniquely defined $\beta_g^{\mathrm{coh}}$ to every row and disables both projection and interface bonding. Active rows comprise all non-residual rows plus residual rows with assigned Gaussians. The initial order is posterior MAP, applicable target-impact, applicable hard-support bonded, support-stiff, solver-compatible, rigid-support, soft, stiff, uncertainty sweep, and alternative material; the first six applicable candidates are retained. Every retained candidate contains every active row. ``Rigid-support'' and ``Repair: rigid support'' are serialized candidate names retained to match the released artifacts; both instantiate the support-shape projection. Deterministic screening diagnostics map support failure to the support repairs, spread or flattening to the cohesion repairs, numerical failure to solver stability, and an otherwise low score to role balancing. At most three unique repairs are emitted in a repair round.

\subsection{Gaussian and Filled-Particle Initialization}

The first $N_G$ particles are initialized at the normalized Gaussian means and retain the corresponding Gaussian indices. Additional particles provide interior support for MPM. Filling is performed within an object-scoped region, so each added particle first receives a source-object identity and only then searches for its nearest Gaussian among that object's Gaussians, as formalized by Equations~(5)--(6) in the main paper. This ordering prevents a filled particle near an object boundary from inheriting the row of a different object. The source Gaussian transfers the persistent object--part row, and the selected candidate row supplies the constitutive branch and continuous coefficients.

Each particle stores its initial position and rest volume $V_q^0$. Candidate-specific mass is formed as $M_q^{(c)}=\rho_q^{(c)}V_q^0$; therefore candidate changes affect material state without changing the shared particle geometry. Gaussian-associated particles are transported back to dynamic 3DGS primitives for rendering, whereas filled particles contribute mass, stress, and momentum during simulation but are not rendered directly. Residual rows ensure that every associated or filled particle receives a valid candidate record.

\subsection{Heterogeneous Step-Size Rule}

Let $\mathfrak R_\Theta$ contain the selected part rows together with the auxiliary whole-object row $(\beta_g,E_g,\nu_g,\rho_g)$ defined above (or its stated fixed fallback), and let $\Delta t_0$ denote the base integration step. For each $a\in\mathfrak R_\Theta$, the compiler forms the elastic wave-speed proxy
\begin{align}
G_a&=\frac{E_a}{2(1+\nu_a)},\notag\\
\lambda_a&=\frac{E_a\nu_a}
{(1+\nu_a)(1-2\nu_a)},\notag\\
c_a&=\sqrt{\frac{\lambda_a+2G_a}{\rho_a}}.
\label{eq:supp-wavespeed}
\end{align}
Let $c_{\max}=\max_{a\in\mathfrak R_\Theta}c_a$, and let $c_{\mathrm{ref}}$ be computed from the nominal tuple $(E,\nu,\rho)=(5{\times}10^6,0.35,2500)$. With the fixed wave-speed scaling coefficient $\kappa_{\mathrm{CFL}}=0.7$, the target integration step is
\begin{equation}
\Delta t_{\mathrm{target}}=
\begin{cases}
\Delta t_0, & c_{\max}\leq \kappa_{\mathrm{CFL}}c_{\mathrm{ref}},\\
\kappa_{\mathrm{CFL}}\Delta t_0\,c_{\mathrm{ref}}/c_{\max},
& c_{\max}>\kappa_{\mathrm{CFL}}c_{\mathrm{ref}}.
\end{cases}
\label{eq:supp-adaptivedt}
\end{equation}
The implementation converts this target into a ceiling-rounded number of substeps and then applies the configured minimum-step and maximum-substep caps. This CFL-inspired wave-speed scaling is the implemented substep-selection heuristic used consistently across candidates.

\subsection{Structure-Preserving Projection}

The main paper gives the complete implemented post-step update. Applied after each output-frame MPM update and before Gaussian transport, the operator acts on support-like rows to reduce local deformation and suppress burr-like rendering artifacts. The optional interface-retention update defined below follows it.

\FloatBarrier

\section{Solver-Aware Skills and Executable Screening}
\label{app:protocol}

\subsection{Supported Scene Actions}

The scene plan represents object roles, action-source/target identities, and the operations in Table~\ref{tab:actions}. The deterministic compiler canonicalizes directions, bounds magnitudes, validates targets, instantiates the actions over explicit spatial regions and intervals, and combines them with the gravity and floor/collider entries supplied by the fixed run specification.

\begin{table}[t]
\centering
{
\footnotesize
\begin{tabularx}{\columnwidth}{@{}p{0.28\columnwidth}X@{}}
\toprule
Action & Implemented operation and spatial scope \\
\midrule
\texttt{translation} & Enforced constant velocity on the selected object or particle region. \\
\texttt{top\_press} & Enforced downward velocity on the selected target's top slab. \\
\shortstack[l]{\texttt{radial\_}\\\texttt{spread}} & Outward horizontal velocity blend plus a bounded position increment on selected deformable particles. \\
\shortstack[l]{\texttt{vertical\_}\\\texttt{compress}} & Position blend toward a reduced vertical extent, with the induced velocity capped, on the selected deformable region. \\
\shortstack[l]{\texttt{side\_}\\\texttt{compress}} & Inward position and velocity blend on the contact-facing slab of the selected target. \\
\texttt{drag\_pair} & Enforced puller velocity; optional serialized flags add interface or whole-target follow velocities. \\
\texttt{scale} & Opposing enforced velocities on two slabs of one object for compression or stretching. \\
\texttt{impulse} & A short particle-force impulse, capped at four base substeps. \\
\texttt{torque} & Bounded tangential particle-velocity field about the selected axis. \\
\shortstack[l]{\texttt{support\_}\\\texttt{hold}} & Zero-velocity boundary condition on an eligible support region. \\
\bottomrule
\end{tabularx}
}
\caption{Supported action vocabulary and its implemented solver operation. ``Enforced velocity'' denotes a bounded particle boundary condition over the action interval; position fields blend the current state toward a locally constructed deformation target. The serialized \texttt{torque} token denotes a rotation request implemented by a tangential velocity field, rather than a directly applied mechanical moment.}
\label{tab:actions}
\end{table}

\subsection{Candidate Budget and Execution Records}

Under the complete Problem Setup contract---motion prompt, four calibrated RGB views and cameras, their aligned static $\mathcal G^0$, and prescribed output cameras---the reported evaluation configuration combines scene-agent-proposed object--part rows and a deterministically validated plan, Qwen3.7-Plus material evidence, and a crop-conditioned feed-forward physical-property distribution for every semantic part. After the Material Reasoning Agent returns row-wise evidence posteriors, the Object-Part Scene Agent invokes candidate construction, retains at most six initial candidates, and controls at most three candidate--execute--observe rounds. Each repair round produces at most three new candidate IDs, and fixed gate/rank rules use acceptance threshold $0.72$. The artifact audit requires a complete, decodable $T$-frame sequence; multi-object rollouts use configurations with interface bonding disabled. Hard-part structure preservation remains candidate-controlled. Each retained artifact contains the object--part records, masks, Gaussian identities, evidence records, complete scene-candidate configurations, execution logs, scores, selected table, runtime, and final flags.

Round 0 evaluates the retained initial candidates; rounds 1 and 2, when reached, evaluate repairs of the current base. \mbox{Candidates} accumulate across rounds, and each candidate ID is evaluated once. At each round, gate-admitted candidates are ranked by $Q_{\mathrm{exec}}$ over the accumulated set, with first appearance in the fixed candidate order breaking ties. If the admitted set is nonempty, the highest-ranked candidate is submitted to the Object-Part Scene Agent for final artifact audit: a successful audit returns that rollout, whereas a failed audit terminates the run with failure, exactly as in Algorithm~1. Only when the admitted set is empty and another round remains does the current highest-ranked executed candidate supply the repair base. Repairs are emitted in this fixed order: support stiffness, rigid support, elastic cohesion, higher cohesion, solver stability, and role balancing; the first three remaining unique IDs are retained. An empty admitted set after the final round returns failure.

\subsection{Initial-Neighborhood Interface Retention}

For a support row $r$, the hard-support-bonded candidate uses support-shape weight $\eta_r=0.85$ and considers particles assigned to a valid compound row other than $r$. It computes each candidate's initial distance to the nearest particle assigned to $r$, keeps distances no greater than $d_b=0.045$ in normalized solver coordinates, orders them by distance, and takes the first 26,000. For each retained particle, it stores $\bm\delta_{qr}^{b,0}=\bm x_q^0-\bar{\bm x}_r^0$. Each frame applies
\begin{align*}
\bm x_q&\leftarrow(1-\eta_b)\bm x_q
+\eta_b(\bar{\bm x}_r+\bm\delta_{qr}^{b,0}),\\
\bm v_q&\leftarrow(1-\zeta_b)\bm v_q
+\zeta_b\bar{\bm v}_r.
\end{align*}
with $\eta_b=\zeta_b=0.95$. Here $\eta_r$ controls the support-shape projection in the main paper, whereas the distinct weights $\eta_b$ and $\zeta_b$ control position and velocity retention at the selected support--part interface. This initial-neighborhood projection preserves selected support--part interfaces during single-object rollouts. Multi-object evaluation uses configurations with interface bonding disabled, retaining object-scoped interaction through the MPM grid.

\subsection{Validity Gate, Ranking, and Tie Rules}

Let $n_{\mathrm{proj}}(k)$ and $n_{\mathrm{bond}}(k)$ be the numbers of rows in candidate $k$ using support-shape projection and interface bonding, respectively. The implementation score $Q_{\mathrm{exec}}$ used to rank rollout candidate $k$ is
\begin{align}
\bm\varphi(k)
&=(s_{\mathrm{vid}}(k),\bar\gamma_k,s_{\mathrm{div}}(k),
s_{\mathrm{role}}(k),s_{\mathrm{base}}(k),\notag\\
&\hspace{20mm}s_{\rho}(k),s_{\mathrm{solver}}(k))^{\mathsf T},\notag\\
\bm w
&=(1,0.27,0.10,0.25,0.15,0.08,0.15)^{\mathsf T},\notag\\
\ell(k)&=\min\!\left(
0.12,0.025n_{\mathrm{proj}}(k)
+0.035n_{\mathrm{bond}}(k)\right),\notag\\
Q_{\mathrm{exec}}(k)
&=\bm w^{\mathsf T}\bm\varphi(k)-\ell(k).
\label{eq:supp-critic}
\end{align}
The video feature $s_{\mathrm{vid}}(k)$ is initialized to zero. A missing video subtracts $0.15$; an existing video smaller than 4096 bytes subtracts $0.20$; and an existing video of at least 4096 bytes adds $0.20$. These terms order executed candidates and repair bases. Independently of this score, the final artifact audit in Algorithm~1 requires the selected artifact to exist, decode successfully, and contain the complete $T$-frame sequence, so no missing or incomplete video can be returned. When valid first/last foreground measurements exist, the score adds $0.12$, then subtracts $0.25$ if foreground-box area grows above $2.0\times$ (or $0.10$ above $1.45\times$), $0.22$ if height falls below $0.45\times$ while width exceeds $1.25\times$, and $0.08$ if the dark-foreground ratio exceeds $0.18$. A nonzero process exit or a numerical/runtime-error token fails the validity gate. Let $\mathcal R_k^{\mathrm{act}}$ be candidate $k$'s active-row set and $N_k^{\mathrm{part}}=|\mathcal R_k^{\mathrm{act}}|>0$. Then $\bar\gamma_k=(N_k^{\mathrm{part}})^{-1}\sum_{r\in\mathcal R_k^{\mathrm{act}}}\gamma_r^{\mathrm{part}}$ averages Equation~\eqref{eq:supp-partconfidence}, and $s_{\mathrm{div}}(k)$ is the number of distinct visual labels divided by $N_k^{\mathrm{part}}$. Candidate construction rejects an empty part table before execution. For every initial candidate, including target-impact, $s_{\mathrm{base}}(k)=\bar\gamma_k$; a repair inherits this value from its base candidate. The serialized base-confidence term is inherited by repairs, while the part-confidence term summarizes aggregated part-level evidence.

The role term uses a separate name-only lookup. A name containing head, blade, tip, plate, support, stand, tray, dish, shell, or frame receives 1 only when raw $E\geq10^6$ Pa and $\beta=\mathrm{Metal}$, and 0.35 otherwise. A name containing rubber, sole, tire, wheel, foam, or cushion receives 1 when raw $E\leq10^8$ Pa and 0.5 otherwise. Every other row receives 0.7. The term $s_{\mathrm{role}}(k)$ averages these values. The density term $s_{\rho}(k)$ averages 0.35 below 250, 0.55 above 5000, and 1 otherwise.

The evaluator uses separate frozen predicates $\widehat S_r(k)$ and $\widehat C_r(k)$. A row satisfies $\widehat S_r(k)=1$ when its name contains plate, dish, tray, bowl, base support, support, stand, holder, case, shell, or frame, or when its visual label is Wood, Metal, Glass, Ceramic, or Stone; it is zero otherwise. It satisfies $\widehat C_r(k)=1$ only when $\widehat S_r(k)=0$, its branch is non-granular, and either the concatenation of its stored name, physics-group token, role, and visual label contains body, base, core, filling, cream, frosting, icing, soft, cushion, pad, foam, sponge, rubber, gel, fruit, or food, or its branch is Foam, Plasticine, or Jelly and its visual label is not one of the five rigid labels above. It is zero otherwise.

For rows with $\widehat C_r(k)=1$, the first matching rule gives the solver score. A Plasticine visual label whose name contains clay, putty, plasticine, or dough receives 1 for Plasticine and 0.7 otherwise. The remaining rows receive 1 for Jelly, 0.55 for Foam or Plasticine, and 0.7 otherwise. The term $s_{\mathrm{solver}}(k)$ averages this lookup over $\widehat C_r(k)=1$ rows and equals 0.7 when that set is empty.

Candidate admission requires a resolved scene plan; successful process termination; no numerical/runtime-error token; foreground-box area growth no greater than $2.0\times$; and no simultaneous height ratio below $0.45\times$ and width ratio above $1.25\times$. In addition, when the set $\{r:\widehat S_r(k)=1\}$ is nonempty, every such support row must have $E_r^{(k)}\geq10^7$ Pa or the candidate must use the solver-compatible regularization transform. This support condition is vacuously satisfied when no support row exists. Candidates satisfying these gates and $Q_{\mathrm{exec}}(k)\geq\tau$, with $\tau=0.72$, are ordered by the same score and then undergo the mandatory artifact audit above. Action references are validated by the scene-plan compiler before rollout. The fixed implementation score provides a consistent criterion for candidate admission and ordering.

\section{Evaluation Protocol}

\subsection{CLIP Similarity and Aggregation}

Let $\phi_I$ and $\phi_T$ be the frozen CLIP image and text encoders, $I_{c,t}$ frame $t$ of case $c$, and $y_c$ its complete motion prompt. Frame similarity and the case score are
\begin{align}
s_{c,t}
&=\frac{\phi_I(I_{c,t})^{\mathsf T}\phi_T(y_c)}
{\|\phi_I(I_{c,t})\|_2\,\|\phi_T(y_c)\|_2},\notag\\
\operatorname{CLIP}_{\mathrm{sim}}(c)
&=\frac{1}{T}\sum_{t=1}^{T}s_{c,t}.
\label{eq:supp-clipsim}
\end{align}
For any reported set, scores are first computed per case and then averaged over its cases. A multi-part case contributes to every material-profile set represented by one of its active semantic rows; the six-profile average is the unweighted mean of those six set scores. Multi-object results are averaged once over the multi-object set without a material subdivision.

\subsection{Reference-Video Motion and Deformation Metrics}
\label{app:reference-metrics}

The evaluation manifest fixes one synthetic MPM reference video for every case. Each reference is generated once from a preregistered, hand-authored MPM configuration using the same static 3DGS, initial pose, camera, frame count, frame time, renderer, and interaction as the compared outputs; its parameters are not changed after method videos have been scored. References are accessed only by the evaluation code---not by material reasoning, candidate construction, screening, or repair. The reported units for all three diagnostics equal 100 times the corresponding raw RMSE. Let $(x_{t,i},y_{t,i})$ and $(\widehat x_{t,i},\widehat y_{t,i})$ denote reference and predicted CoTracker3 coordinates for point $i$ at frame $t$. With frame width $W$, height $H$, and $\mathcal J$ the point--frame pairs visible in both videos, reported trajectory nRMSE is
\begin{equation}
\begin{aligned}
E_{\mathrm{traj}}
&=100\Biggl[\frac{1}{|\mathcal J|}
\sum_{(t,i)\in\mathcal J}\Biggl(
\frac{(\widehat x_{t,i}-x_{t,i})^2}{W^2}\\[-0.2ex]
&\hspace{7.5em}
+\frac{(\widehat y_{t,i}-y_{t,i})^2}{H^2}
\Biggr)\Biggr]^{1/2}.
\end{aligned}
\label{eq:supp-traj}
\end{equation}
The reference is tracked once from a fixed $16\times16$ query grid restricted to its eroded frame-0 target mask, and every method reuses those same frame-0 queries. A pair belongs to $\mathcal J$ only when the same initialized point is visible, remains inside the registered target mask, and has a finite track in both videos at that frame. Frames with fewer than 16 such points are excluded. Restricting the sum to jointly visible pairs prevents a missing measurement from being treated as zero displacement. The metric record is valid only when at least one frame remains; otherwise the output is recorded as invalid rather than assigned zero error.

\begin{samepage}
The target masks are evaluation-only visible-target silhouettes rendered with all scene Gaussians retained as occluders. For target-mask areas $A_t$ and $\widehat A_t$, we compute projected-area log-RMSE as
\begin{equation}
E_{\mathrm{area}}=
100\left[\frac{1}{T}\sum_{t=1}^{T}
\bigl(\log(\widehat A_t+1)-\log(A_t+1)\bigr)^2
\right]^{1/2}.
\label{eq:supp-area}
\end{equation}
\end{samepage}
The logarithm measures relative area change; the $+1$ term keeps an associated empty mask finite, while failed target association yields an invalid record.

For local deformation, each frame-0 tracked point uses its eight nearest tracked neighbors. A least-squares local 2D deformation gradient $\bm F_{t,i}$ maps the initial neighborhood offsets to their offsets at frame $t$, giving Green--Lagrange strain
\begin{equation}
\bm E_{t,i}=\tfrac12
\bigl(\bm F_{t,i}^{\mathsf T}\bm F_{t,i}-\bm I_2\bigr).
\label{eq:supp-strain}
\end{equation}
The reference-defined eight-neighbor graph is reused for every method. The set $\mathcal K$ contains only point--frame pairs for which the center and at least four of its eight stored neighbors are jointly visible with finite tracks and the regularized local least-squares system is nondegenerate. For this jointly valid set, local-strain RMSE is
\begin{equation}
E_{\mathrm{strain}}=
100\left[\frac{1}{|\mathcal K|}
\sum_{(t,i)\in\mathcal K}
\|\widehat{\bm E}_{t,i}-\bm E_{t,i}\|_F^2
\right]^{1/2}.
\label{eq:supp-strain-error}
\end{equation}
This removes rigid translation while retaining local stretch, compression, and shear discrepancies. The local-strain record is valid only when $|\mathcal K|>0$; an empty valid-neighborhood set marks the output invalid rather than assigning it zero. Invalid runs remain in the evaluation registry as failures rather than receiving an artificial zero error. The finalizer aborts rather than dropping or imputing any required invalid record; the reported reference-metric tables were finalized only after their registries contained zero invalid records. Multi-part results are averaged within each applicable profile and then macro-averaged over the six profiles. For multi-object motion, registered objects are evaluated separately, averaged with equal weight within a scene, and then averaged across scenes. Together, the three diagnostics quantify image-plane trajectory, projected deformation scale, and local strain relative to synchronized reference videos.

\subsection{User Preference Evaluation}
\label{app:userstudy}

\paragraph{Study design.}
The reported UPR aggregates responses from 20 researchers with Computer Vision or Computer Graphics expertise. The main comparison uses a Five-Alternative Forced-Choice (5AFC) study: each trial presents synchronized outputs from OmniPhysGS, PhysSplat, DreamPhysics, PhysGM, and \method{}. The ablation uses a separate Three-Alternative Forced-Choice (3AFC) study containing the two ablated variants and the full model. Method names are hidden, and the videos are shown in randomized positions. For every case, a respondent selects exactly one video that best matches the motion prompt while exhibiting physically plausible and visually natural dynamics. Every case remains in the study through an explicit output-status card where applicable. The 5AFC and 3AFC protocols are reported separately according to their respective choice sets and chance levels.

\paragraph{Aggregation.}
Let $\mathcal S$ be a reported set of evaluation cases, $R_c$ the number of valid responses for case $c$, and $J_{ch}$ the method selected by respondent $h$ for that case. The per-case preference share and the reported User Preference Rate are
\begin{align}
u_c(m)
&=\frac{100}{R_c}\sum_{h=1}^{R_c}
\mathbf 1[J_{ch}=m],\notag\\
\operatorname{UPR}_{\mathcal S}(m)
&=\frac{1}{|\mathcal S|}
\sum_{c\in\mathcal S}u_c(m).
\label{eq:supp-upr}
\end{align}
Thus every case receives equal weight even if valid-response counts differ. For the multi-object evaluation, $\mathcal S$ contains all evaluated multi-object cases. For a material-profile column, $\mathcal S$ contains every evaluated multi-part case carrying that profile; a case with several profiles contributes independently to each applicable set. The six-profile average is the unweighted mean of the six profile-level UPR values. UPR reports human preference for prompt consistency, physically plausible motion, and visual naturalness under the stated choice protocol.

\section{Limitations and Future Work}

\paragraph{Cross-View Grounding and Scene Inputs.}
Future work can strengthen persistent object--part assignment through occlusion-aware cross-view association, confidence-calibrated residual handling, and additional observations.

\paragraph{Constitutive and Interaction Coverage.}
New material branches, joints, and topology-aware updates can extend the six-profile interface to a broader range of part behaviors and interactions.

\paragraph{Spatial and Temporal Resolution.}
Adaptive particle/grid resolution and temporal planning can support finer interactions and longer action schedules while preserving the current agent--compiler interface.

\section{Additional Results}
\label{app:additional-results}

\begin{figure*}[p]
\centering
\includegraphics[width=\textwidth,height=0.82\textheight,keepaspectratio]{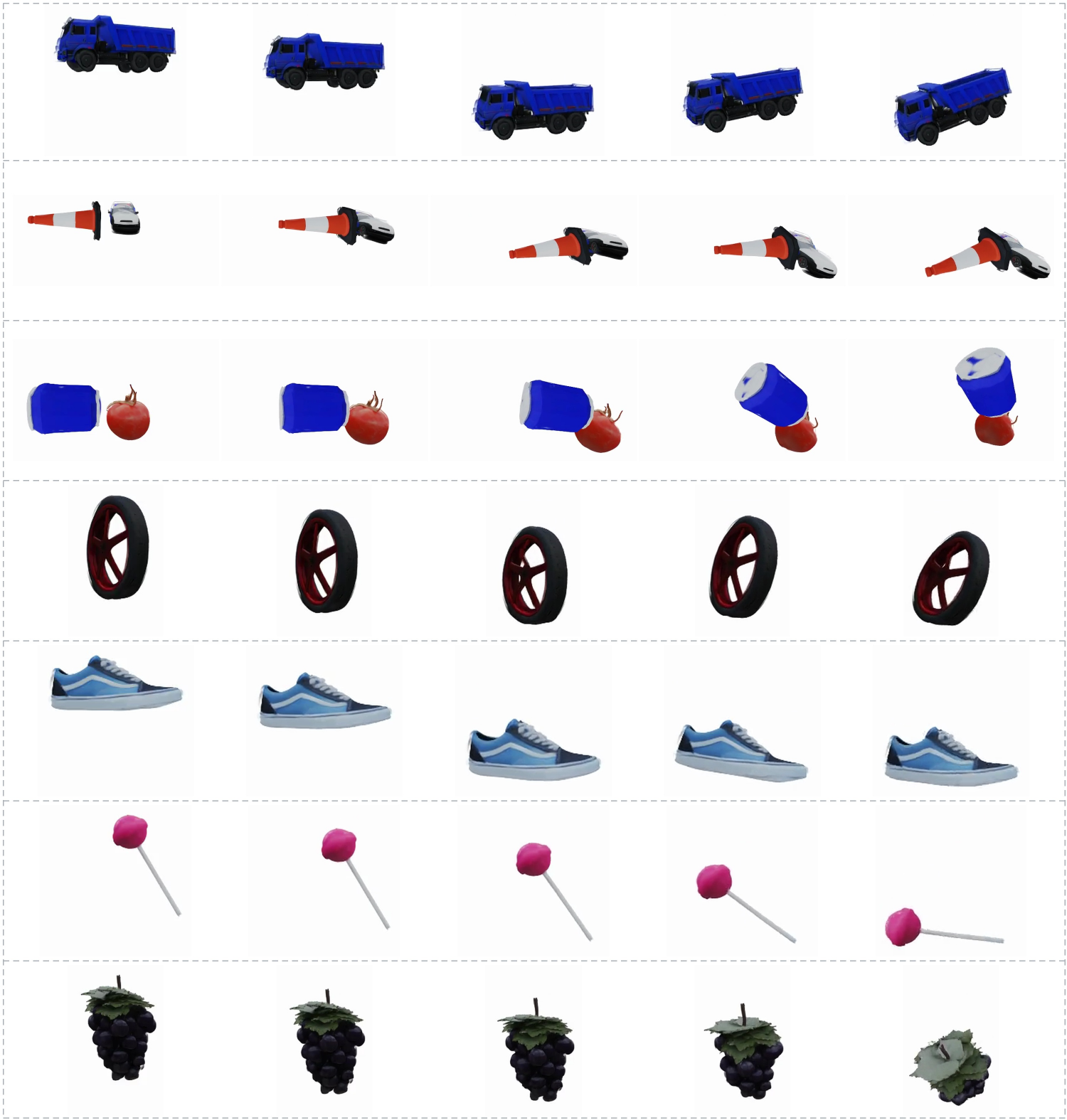}
\caption{Additional results (I).}
\label{fig:additional-gallery-1}
\end{figure*}

\begin{figure*}[p]
\centering
\includegraphics[width=\textwidth,height=0.82\textheight,keepaspectratio]{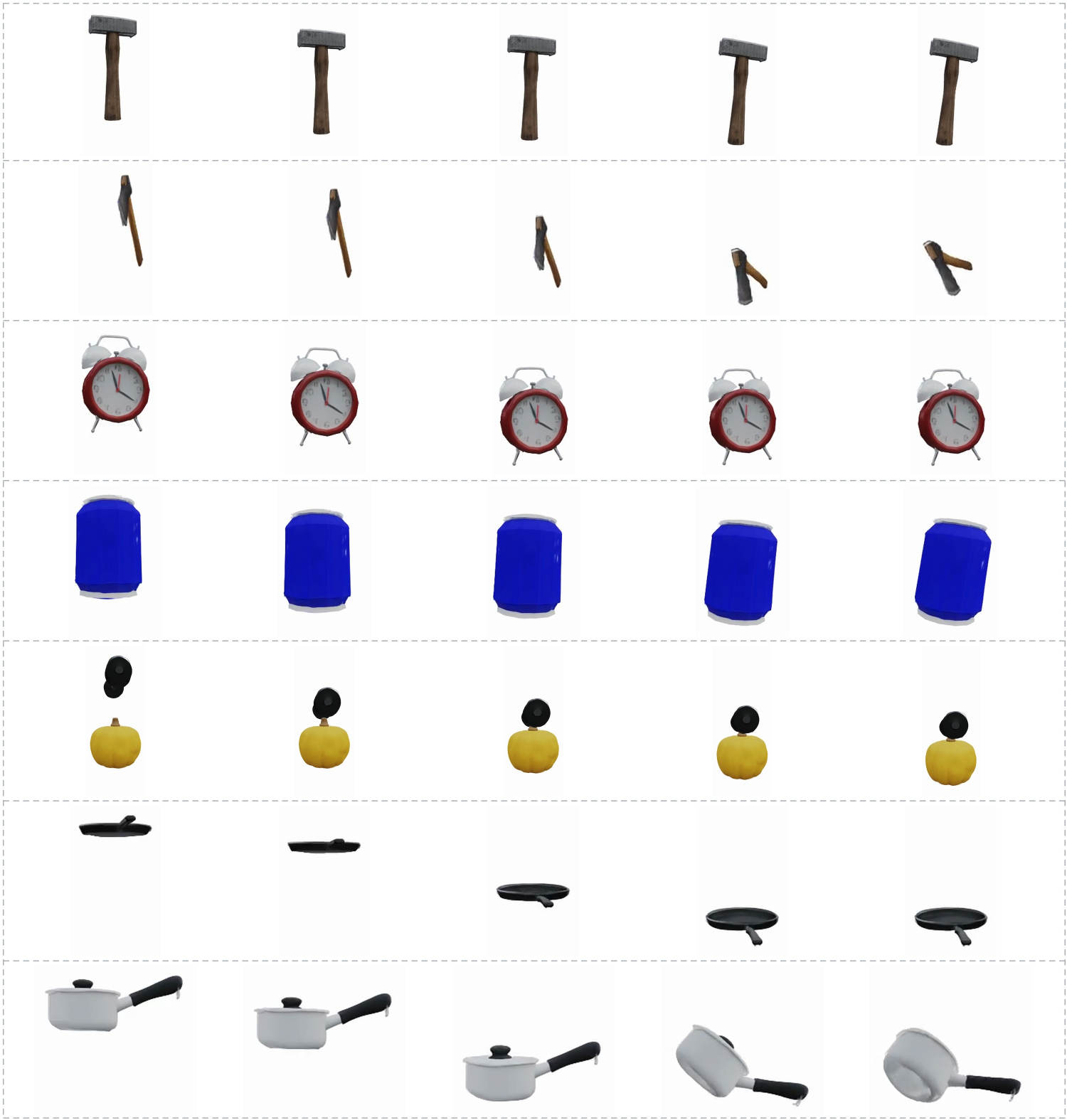}
\caption{Additional results (II).}
\label{fig:additional-gallery-2}
\end{figure*}

\begin{figure*}[p]
\centering
\includegraphics[width=\textwidth,height=0.82\textheight,keepaspectratio]{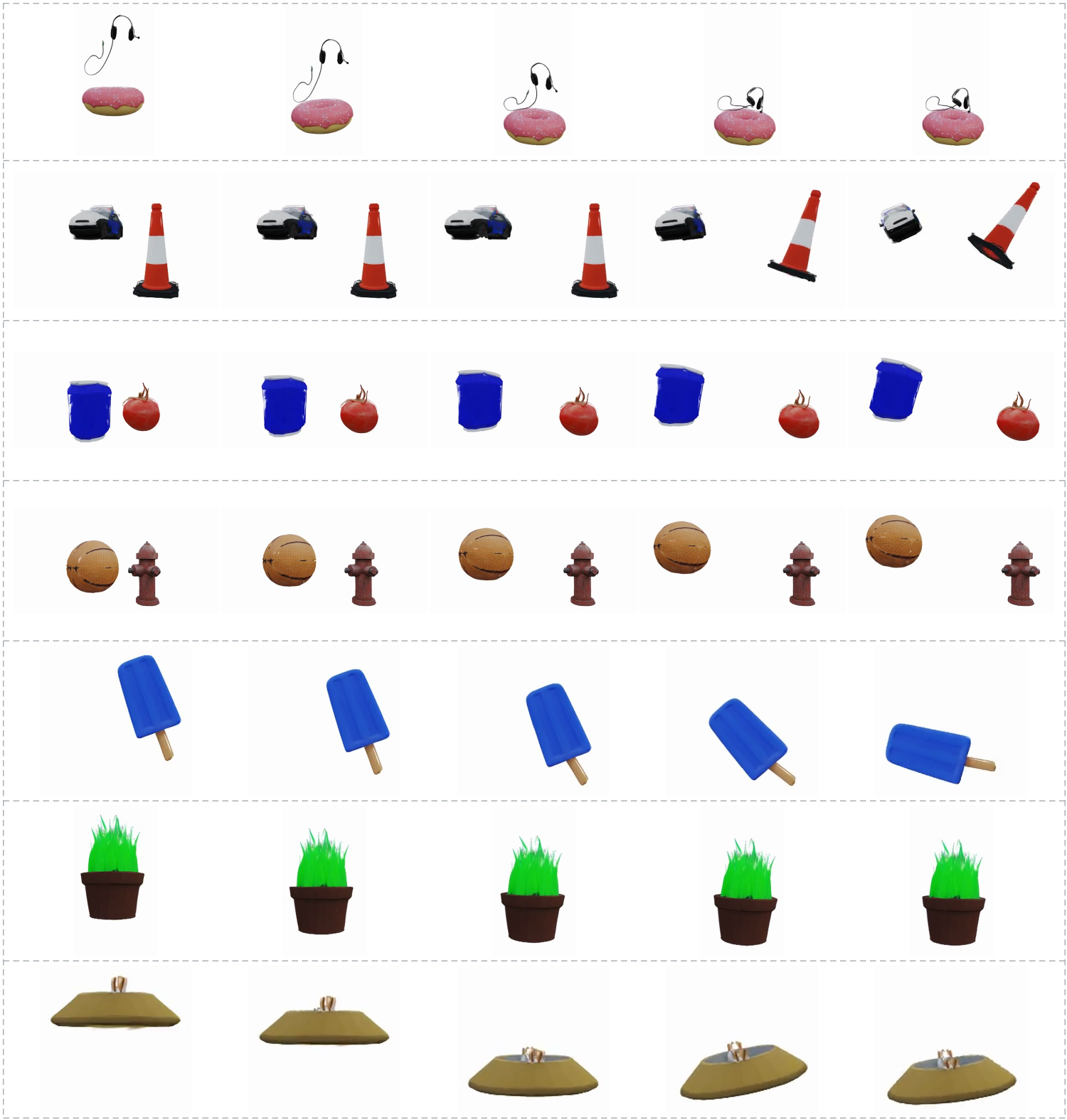}
\caption{Additional results (III).}
\label{fig:additional-gallery-3}
\end{figure*}

\begin{figure*}[p]
\centering
\includegraphics[width=\textwidth,height=0.82\textheight,keepaspectratio]{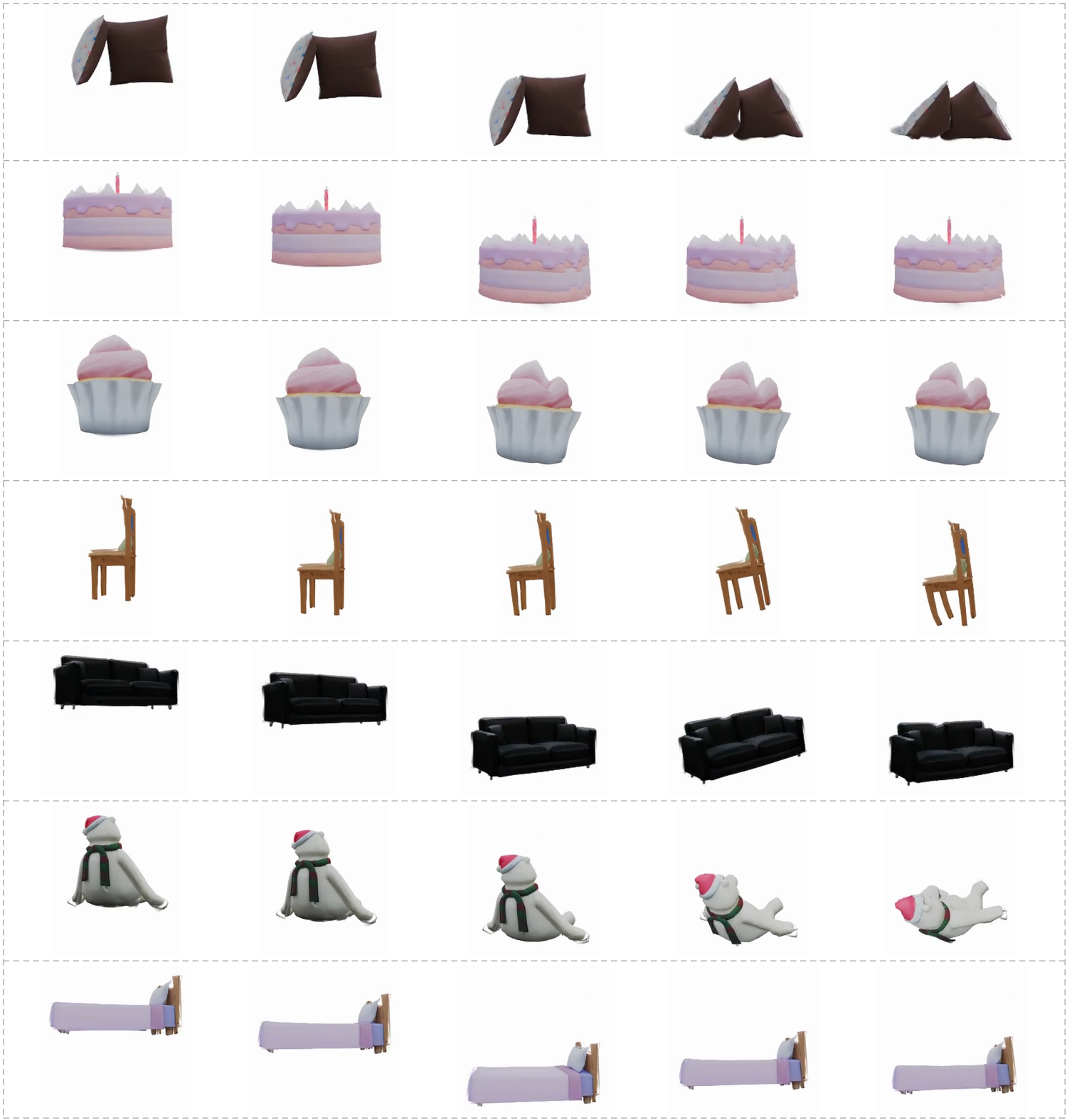}
\caption{Additional results (IV).}
\label{fig:additional-gallery-4}
\end{figure*}

\begin{figure*}[p]
\centering
\includegraphics[width=\textwidth,height=0.82\textheight,keepaspectratio]{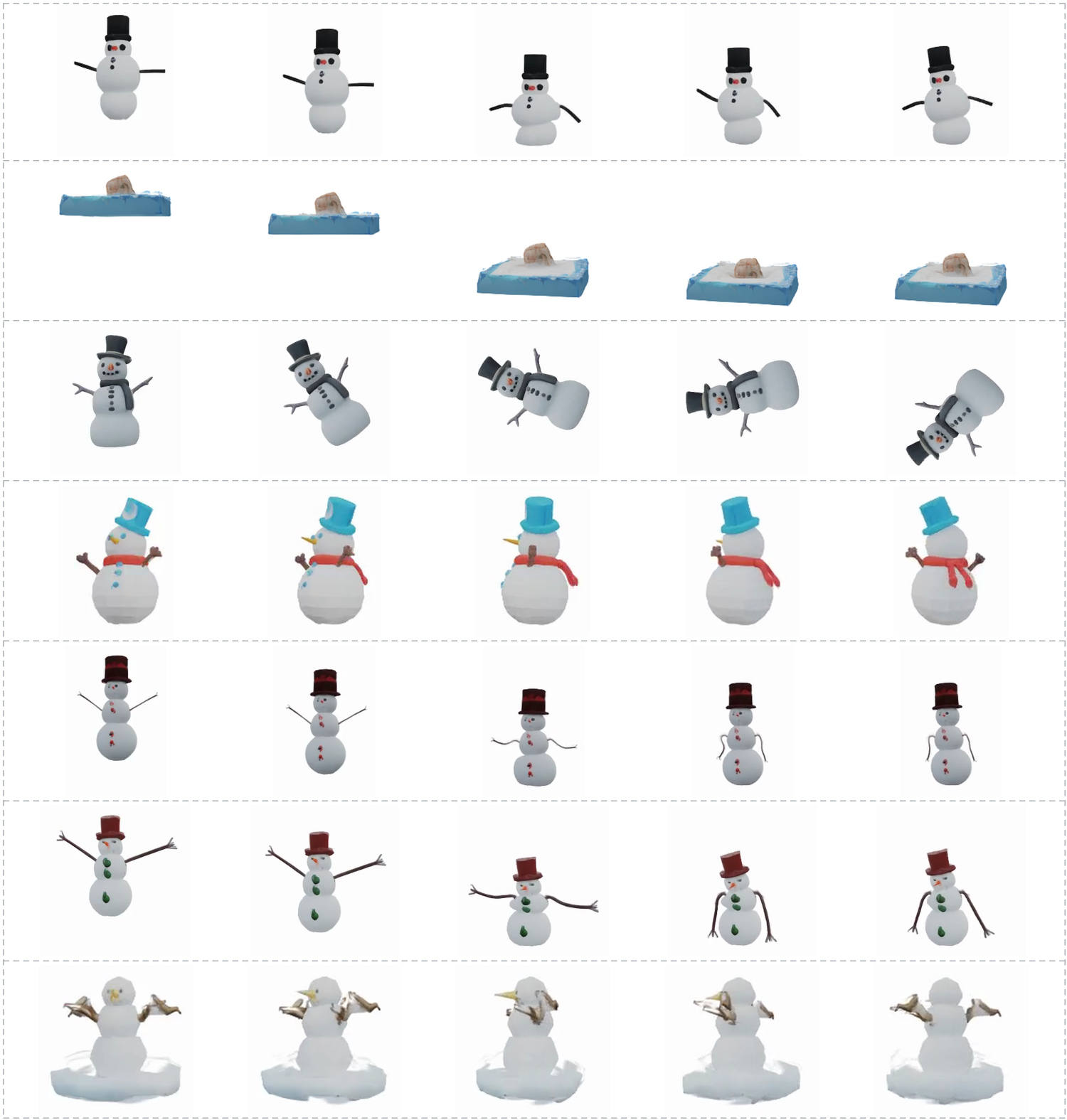}
\caption{Additional results (V).}
\label{fig:additional-gallery-5}
\end{figure*}

\begin{figure*}[p]
\centering
\includegraphics[width=\textwidth,height=0.82\textheight,keepaspectratio]{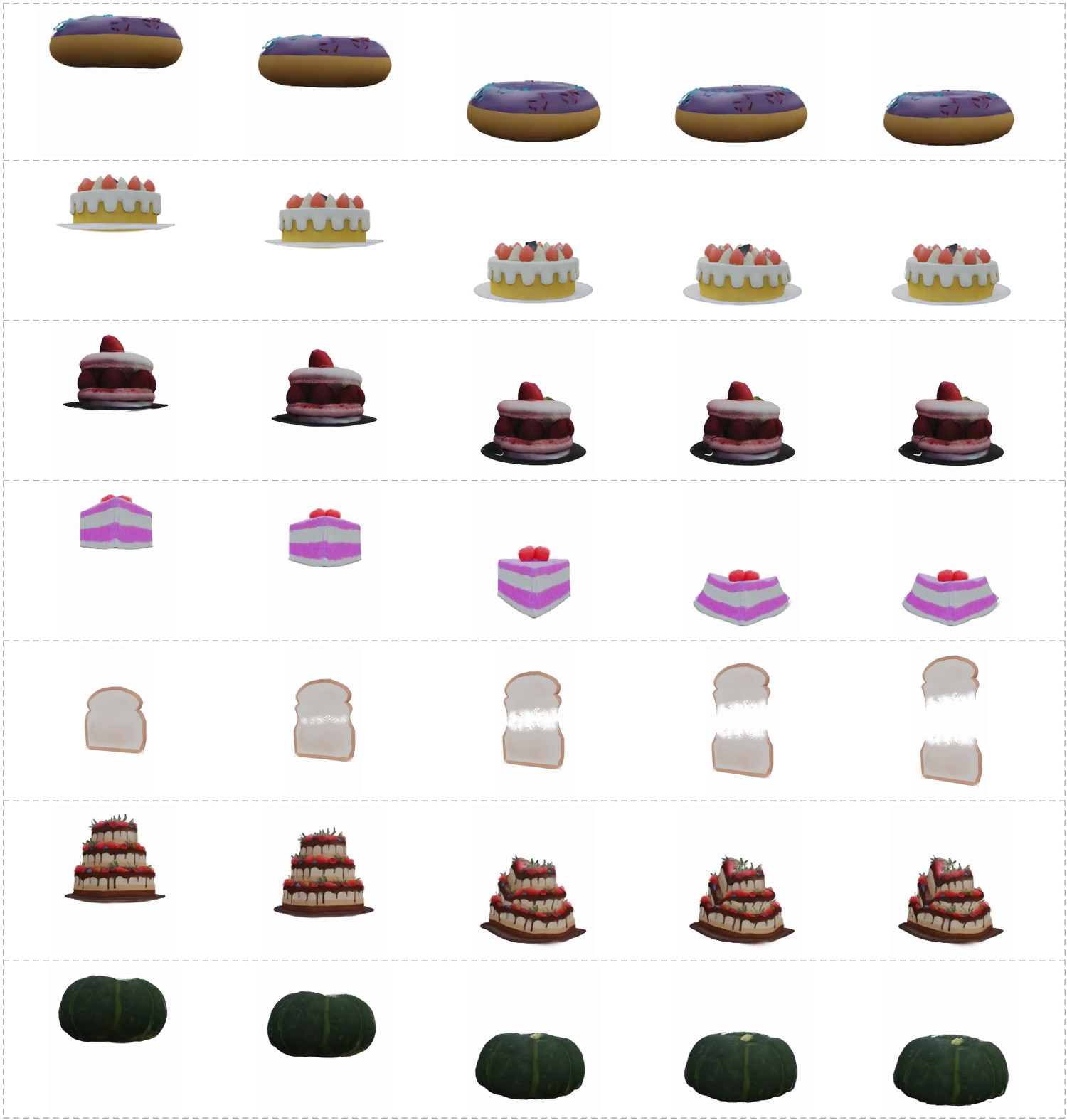}
\caption{Additional results (VI).}
\label{fig:additional-gallery-6}
\end{figure*}

Figures~\ref{fig:additional-gallery-1}--\ref{fig:additional-gallery-6} present 42 additional \method{} rollouts. Each row shows five temporally ordered frames, covering heterogeneous multi-part objects and multi-object scenes under varied motion configurations.

\FloatBarrier

\end{document}